\documentclass[runningheads]{llncs}

\usepackage{accv}

\usepackage{accvabbrv}

\usepackage{graphicx}
\usepackage{booktabs}
\usepackage{array}
\usepackage{multirow}
\usepackage{xcolor}
\usepackage{tabularx}
\usepackage{enumitem}
\usepackage{placeins}

\usepackage[accsupp]{axessibility}  

\usepackage{hyperref}
\usepackage[capitalize]{cleveref}

\usepackage{orcidlink}

\begin{document}

\title{CASE: Concept-Aware Segmentation from Positive and Negative Visual Exemplars} 

\titlerunning{CASE: Concept-Aware Segmentation from Exemplars}

\author{Yi-Jen Tsai\inst{1} \and
Yen-Yu Lin\inst{1} \and
Chien-Yao Wang\inst{2}\thanks{Corresponding author}}

\authorrunning{Y.-J. Tsai et al.}

\institute{National Yang Ming Chiao Tung University, Taiwan\\
\email{tsai.cs14@nycu.edu.tw, lin@cs.nycu.edu.tw} \and
Institute of Information Science, Academia Sinica, Taiwan\\
\email{kinyiu@iis.sinica.edu.tw}}

\maketitle

\vspace{-4mm}
\begin{abstract}
Few-shot semantic segmentation typically identifies target objects based on positive support examples of a single semantic class. In practice, however, positive cues alone are often underspecified: for instance, ``red cars'' may be confused with cars of all colors, ``vehicles'' may include cars that should be excluded, and ``material-defined objects'' may span multiple categories. 
Fulfilling precise requests requires positive exemplars to provide inclusion evidence and negative exemplars to act as intent-defining constraints for exclusion and disambiguation. We formulate this setting as \textbf{Specified Visual Prompt Segmentation (SVPS)}, where the target concept is specified by both positive and negative visual prompts. 
In this work, visual exemplars refer to support-image regions specified by bounding-box prompts. To study SVPS, we introduce \textbf{CASE} (\textbf{C}oncept-\textbf{A}ware \textbf{S}egmentation from Positive and Negative Visual \textbf{E}xemplars), a benchmark built on COCO and PACO. CASE encompasses six category- and attribute-level task families: exclusion, isolation, union, attribute exclusion, attribute isolation, and attribute union. 
We evaluate SAM 3 under this setting and find that, despite its powerful positive-only few-shot segmentation capability, frozen SAM 3 does not consistently treat negative visual prompts as exclusion evidence. To address this, we present \textbf{Concept Prompt Adaptation (CPA)}, a conservative LoRA-based adaptation baseline that improves positive-negative prompt interpretation on CASE while remaining competitive under positive-only few-shot segmentation. Code is available at \url{https://github.com/WongKinYiu/FSS-CASE}.

\end{abstract}

\begin{figure*}[t]
\centering
\includegraphics[width=0.85\linewidth]{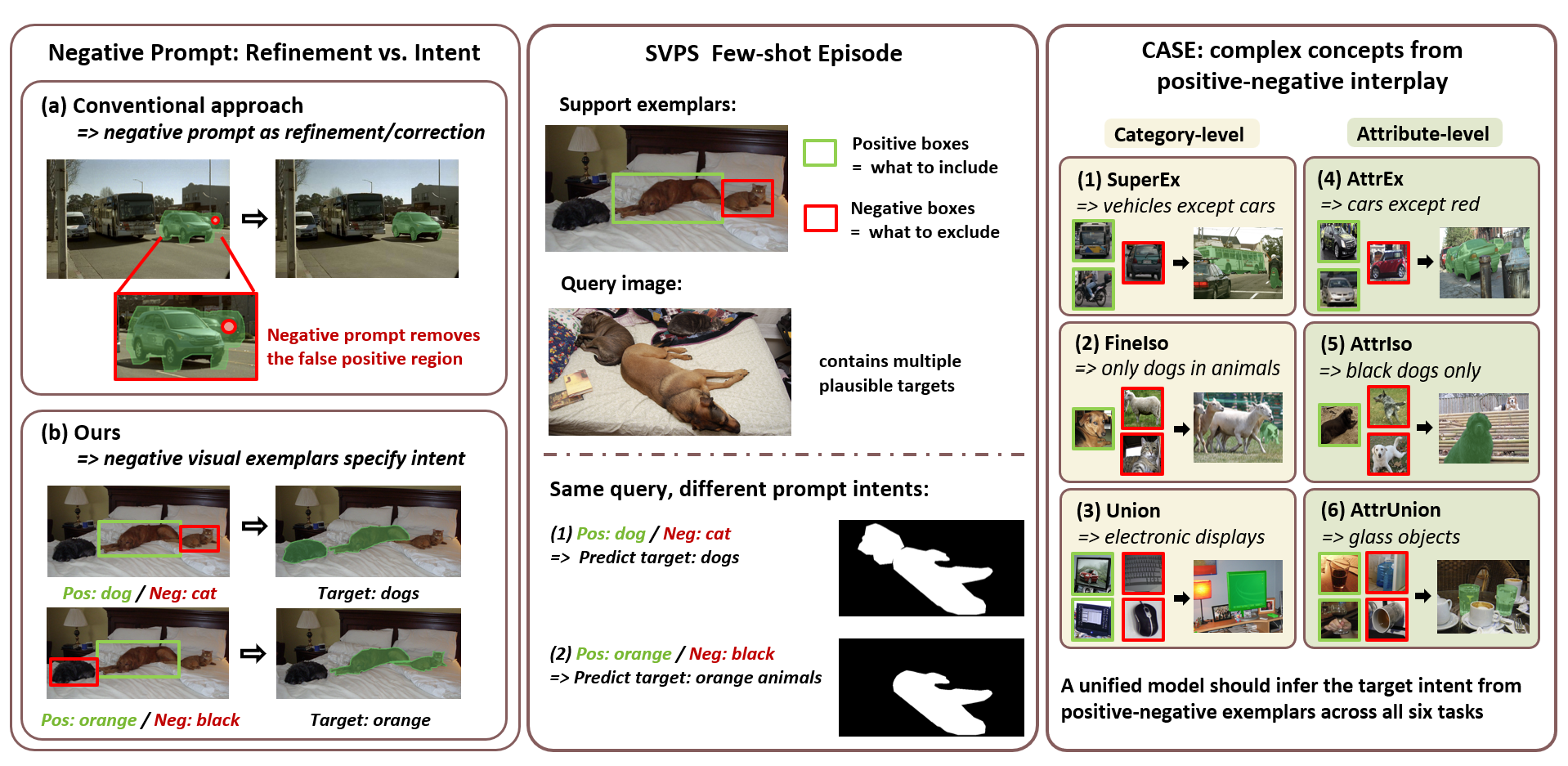}
\vspace{-2mm}
\caption{
\textbf{Motivation and overview of CASE.}
Conventional negative prompts are commonly used as local refinement signals for an already specified target.
In contrast, Specified Visual Prompt Segmentation (SVPS) treats positive and negative visual exemplars as intent-defining cues.
The same query image can therefore induce different target under different prompts.
CASE evaluates this setting through six task families, requiring a unified model to infer target intent from positive-negative exemplar.
}
\label{fig:teaser}
\vspace{-4mm}
\end{figure*}

\vspace{-4mm}
\section{Introduction}
\label{sec:intro}

Semantic segmentation~\cite{long2015fcn, zhao2017pspnet, chen2018deeplab} is a fundamental task in computer vision that aims to assign a semantic label to every pixel in an image. Although deep models have achieved remarkable performance on standard benchmarks, they rely on large-scale datasets with dense, pixel-level annotations for training. Beyond expensive annotation costs, most models are trained on a fixed set of categories, resulting in limited generalization to unseen objects in open-world scenarios.


Few-shot semantic segmentation (FSS)~\cite{wang2019panet,tian2020pfenet} addresses this limitation by segmenting novel objects in a query image using only a small number of annotated support examples. Existing FSS methods commonly adopt episodic training to learn transferable matching mechanisms between support and query images, such as prototype matching~\cite{wang2019panet,tian2020pfenet}, pixel-level correspondence~\cite{min2021hsnet,peng2023hdmnet}, or cross-image attention~\cite{zhang2022fptrans,xu2023sccan}. Despite their success on standard FSS benchmarks, these methods largely formulate the problem as category-level matching: Namely, given one or more positive support examples, the model is expected to segment all objects of the same semantic class in the query image.


However, this formulation is overly restrictive for practical visual prompting. In real-world scenarios, user specifications frequently require both inclusion and exclusion cues, rather than purely positive category matching. For instance, positive examples may overgeneralize from ``red cars'' to all cars, from ``buses and trains'' to all vehicles (thereby incorrectly including cars), or from ``wooden chairs'' to all chairs rather than wooden objects. These complex targets cannot be precisely encapsulated by positive examples alone: While positive examples define what to include, they do not explicitly specify what should be excluded. Consequently, a model needs to jointly interpret positive and negative visual prompts to infer the target mask from their relationship.

We define this task as \textbf{Specified Visual Prompt Segmentation (SVPS)}, where the target concept is specified by both positive and negative visual exemplars. In this work, visual exemplars are support-image regions represented as bounding-box prompts. Positive exemplars provide visual evidence for inclusion, whereas negative exemplars act as intent-defining constraints that specify confusing categories, attributes, or instances to exclude. Unlike conventional FSS, SVPS evaluates whether a model can infer a specified target concept from the interplay between positive and negative exemplars. As illustrated in \cref{fig:teaser}, this changes the role of negative prompts from local mask refinement to concept-level intent specification: different positive-negative exemplar configurations can induce different target masks for the same query image.


To evaluate this task, we introduce \textbf{CASE} (\textbf{C}oncept-\textbf{A}ware \textbf{S}egmentation from Positive and Negative Visual \textbf{E}xemplars), a benchmark constructed from COCO~\cite{lin2014coco} and PACO~\cite{ramanathan2023paco} annotations. The CASE benchmark encompasses six task families spanning category-level and attribute-level concept specification: exclusion, isolation, and union. These tasks are designed to assess whether a segmentation model can leverage negative visual prompts as concept-level exclusion evidence, rather than simply appending them as additional prompts without modeling their relation to the positive visual evidence.


We further study Segment Anything Model 3 (SAM 3)~\cite{carion2026sam3} under the SVPS setting. SAM 3 is a strong vision foundation model for promptable concept segmentation, supporting segmentation from text descriptions, visual exemplars, or their combination. A recent shared-canvas formulation~\cite{tsai2026fsssam3} has shown that support and query images can be placed in a single spatial canvas, enabling a fully frozen SAM 3 to perform few-shot segmentation without additional training or architectural modification. This makes SAM 3 a suitable testbed for studying whether promptable segmentation models can handle positive-negative visual prompts beyond positive-only matching. Unlike general promptable concept segmentation in SAM 3~\cite{carion2026sam3}, where different prompt types serve as alternative ways to refer to a concept, SVPS evaluates whether positive and negative exemplars can jointly specify a constrained target concept.


However, the same line of evidence also reveals a key limitation: directly adding negative visual prompts to the few-shot segmentation pipeline does not necessarily improve performance and may even suppress target objects~\cite{tsai2026fsssam3}. This behavior contradicts the intended role of negative prompts as exclusion cues. It suggests that, although SAM 3 can effectively exploit positive visual prompts, it does not inherently parse negative prompts as precise exclusion constraints. In few-shot settings, negative prompts may instead compete with positive prompts and weaken or suppress predicted target masks.


To address this issue, we propose \textbf{Concept Prompt Adaptation (CPA)}, a lightweight, LoRA-based adaptation strategy tailored for SAM 3 under positive-negative visual prompting. We use the term \emph{concept} to highlight that CPA adapts the model to target concepts specified by positive-negative visual evidence. CPA updates only low-rank adaptation parameters while keeping the pretrained model frozen. Furthermore, we introduce positive-only preservation regularization that keeps the adapted model close to frozen SAM 3 when negative prompts are absent. This provides a conservative yet effective adaptation baseline that balances SVPS improvement on CASE while mitigating standard FSS degradation.

The primary contributions of this work are summarized as follows:
\begin{itemize}[noitemsep, topsep=0pt, partopsep=0pt]
    \item We introduce \textbf{SVPS}, a new segmentation paradigm where positive and negative visual exemplars jointly specify a target concept.
    \item We present \textbf{CASE}, a benchmark for SVPS built upon COCO and PACO, covering six category- and attribute-level concept specification tasks.
    \item We propose \textbf{CPA}, a lightweight LoRA-based adaptation baseline equipped with positive-only preservation regularization, and evaluate the adaptation-preservation trade-off on both SVPS and positive-only FSS benchmarks.
\end{itemize}

\section{Related Work}
\label{sec:related}

\subsection{Few-Shot Semantic Segmentation}

Few-shot semantic segmentation (FSS)~\cite{wang2019panet,tian2020pfenet} aims to segment novel categories in a query image given only a small number of annotated support examples. Early approaches commonly adopt episodic training to learn matching functions. A prominent line of work represents support evidence with prototypes or prototype-enhanced class summaries~\cite{wang2019panet,tian2020pfenet,lang2022bam,fan2022ssp}. Although prototype-based methods are efficient, compressing multiple support masks into a global vector discards fine-grained spatial details crucial for precise localization.


To preserve richer support-query correspondences, subsequent methods explore pixel- or region-level matching through hypercorrelation~\cite{min2021hsnet}, dense matching~\cite{peng2023hdmnet}, or cross-attention~\cite{zhang2022fptrans,xu2023sccan}. Despite effectiveness on FSS, these methods work under the restrictive assumption that the support set provides solely positive evidence for a single semantic class. Hence, the task is reduced to category-level matching, where the model merely transfers a positive support concept to the query image. In contrast, our work studies a more expressive setting where a target concept is jointly specified by positive and negative visual prompts.


\vspace{-2mm}
\subsection{Foundation Models for Promptable Segmentation}

Vision foundation models such as CLIP~\cite{radford2021clip} and DINO~\cite{caron2021dino,oquab2023dinov2} provide general-purpose visual representations that have reshaped many recognition~\cite{zhou2022coop,zhou2022cocoop,zhang2022tipadapter} and segmentation~\cite{rao2022denseclip,zhou2022maskclip,luddecke2022clipseg,xu2022groupvit} tasks. Building on this trend, the Segment Anything Model (SAM)~\cite{kirillov2023sam} demonstrates remarkable class-agnostic segmentation capabilities through spatial prompts such as points and boxes. Several works adapt SAM to FSS by generating visual reference prompts from support images~\cite{sun2024vrpsam,zhang2024gfsam,liu2024matcher}. While effective, these methods often introduce additional components, such as visual reference encoders~\cite{sun2024vrpsam}, graph-based prompt refinement modules~\cite{zhang2024gfsam}, or prompt-learning/adaptation modules~\cite{chen2023samadapter,huang2024ssprompt}, to bridge the gap between support examples and query segmentation.


SAM 2~\cite{ravi2025sam2} extends promptable segmentation to videos through a memory mechanism for frame-to-frame mask propagation. Recent FSS methods based on SAM 2, such as FSSAM~\cite{xu2025fssam} and SANSA~\cite{cuttano2025sansa}, adapt this memory mechanism by treating support images as memory references and query images as target frames. However, these memory-based formulations are primarily designed for temporal consistency and object tracking, making them suboptimal for semantic correspondences across temporally disjoint images.


SAM 3~\cite{carion2026sam3} introduces Promptable Concept Segmentation (PCS), where an object concept can be specified by text prompts, positive or negative image exemplars, or their combinations. Native prompting operates on the image being segmented, whereas CASE requires cross-image transfer. A recent \emph{shared-canvas} pipeline~\cite{tsai2026fsssam3} further enables frozen SAM 3 to perform training-free FSS by placing support and query images into a single canvas, but directly incorporating negative prompts can degrade segmentation quality and suppress target objects. Rather than treating negative exemplars as interactive refinements to an existing concept prompt, SVPS studies whether positive and negative visual exemplars can jointly define complex target concepts through inclusion and exclusion roles.

\vspace{-2mm}
\subsection{Visual Prompting and Negative Prompts}

Prompt design is crucial to foundation segmentation models. Existing works mainly improve positive guidance through visual reference prompts~\cite{zhang2024persam,sun2024vrpsam}, support-query feature matching~\cite{liu2024matcher,zhang2024gfsam}, learnable prompts or adaptation modules~\cite{chen2023samadapter,huang2024ssprompt}, and text- or image-guided prompting~\cite{luddecke2022clipseg,wang2022cris,ding2022lavt}. These mechanisms are used to localize a target object or transfer a positive support concept to a query image. Target concept ambiguity is commonly mitigated by strengthening positive evidence through refined localization~\cite{zhang2024persam}, support-query matching~\cite{liu2024matcher,zhang2024gfsam}, or multimodal guidance~\cite{luddecke2022clipseg,wang2022cris,ding2022lavt,rao2022denseclip}.


Negative prompts remain less explored in exemplar-based FSS. While negative points or boxes are commonly expected to suppress unwanted regions in user-interactive segmentation~\cite{kirillov2023sam,ravi2025sam2}, this assumption may not transfer directly to FSS, where negative exemplars define semantic exclusion constraints rather than merely correct local mask errors. Recent SAM 3-based FSS results~\cite{tsai2026fsssam3} suggest that directly appending negative prompts reduces segmentation quality and suppresses target objects. This indicates that current promptable segmentation models fail to utilize negative visual prompts as exclusion evidence.


Beyond category-level segmentation, referring expression segmentation specifies targets through language descriptions involving attributes or context~\cite{kazemzadeh2014referitgame,yu2016modelingcontext,wang2022cris,ding2022lavt}, while datasets such as PACO~\cite{ramanathan2023paco} evaluate attribute-level visual understanding. Although language can describe constraints, subtle visual distinctions are easier to specify through visual examples.
%
Motivated by this gap, we define Specified Visual Prompt Segmentation (\textbf{SVPS}), where positive and negative visual prompts jointly specify a target concept rather than serving as auxiliary inputs. The proposed \textbf{CASE} (\textbf{C}oncept-\textbf{A}ware \textbf{S}egmentation from Positive and Negative Visual \textbf{E}xemplars) benchmark systematically evaluates whether models can infer target concepts from positive-negative exemplar relationships, covering exclusion, isolation, union, and attribute-level disambiguation.

\section{The CASE Benchmark}
\label{sec:benchmark}

\begin{table*}[t]
\vspace{-2mm}
\centering
\caption{\textbf{Task taxonomy of the CASE benchmark.} Each task specifies a target concept through positive and negative visual exemplars.}
\label{tab:svps_tasks}
\tiny
\setlength{\tabcolsep}{2.5pt}
\renewcommand{\arraystretch}{1.08}
\begin{tabularx}{\textwidth}{@{}p{1.15cm}
>{\raggedright\arraybackslash}X
>{\raggedright\arraybackslash}X
>{\raggedright\arraybackslash}X
>{\raggedright\arraybackslash}X@{}}
\toprule
Task & Target Concept & Positive Exemplars & Negative Exemplars & Challenge \\
\midrule

SuperEx &
Supercategory excluding one class, e.g., vehicles except cars. &
Included classes from the supercategory, e.g., buses and trains. &
The excluded class, e.g., cars. &
Generalize to a broad group while enforcing explicit exclusion. \\

FineIso &
One fine-grained class within a supercategory, e.g., dogs. &
Target class instances, e.g., dogs. &
Related distractor classes, e.g., cats or sheep. &
Isolate a target class from visually close distractors. \\

Union &
Classes sharing a common function, e.g., furniture for resting. &
Functionally related classes, e.g., beds and sofas. &
Related classes with different functions, e.g., chairs or tables. &
Infer functional commonality across distinct objects. \\

AttrEx &
A category excluding one attribute, e.g., non-red cars. &
Instances without the excluded attribute, e.g., black or white cars. &
Instances with the excluded attribute, e.g., red cars. &
Apply visual negation by fine-grained attribute differences. \\

AttrIso &
A category with one specified attribute, e.g., red cars. &
Target attribute instances, e.g., red cars. &
Same category with other attributes, e.g., blue or white cars. &
Separate attribute- specific targets from general object identity. \\

AttrUnion &
An attribute shared across categories, e.g., wooden objects. &
Diverse classes with the attribute, e.g., wooden chairs or bowls. &
Similar classes without the attribute, e.g., plastic chairs or bowls. &
Group objects by attribute while avoiding category or shape bias. \\
\bottomrule
\end{tabularx}
\vspace{-4mm}
\end{table*}

\subsection{Benchmark Overview}

We introduce \textbf{CASE}, a benchmark for \textbf{SVPS}. Traditional few-shot segmentation benchmarks evaluate whether a model can transfer a single positive support concept to a query image. In contrast, CASE requires models to infer target concepts by joint reasoning over both positive and negative visual exemplars. Positive exemplars define regions to include, whereas negative exemplars specify related categories, attributes, or instances that must be excluded.


CASE is built on COCO~\cite{lin2014coco} and PACO~\cite{ramanathan2023paco}. COCO provides category and supercategory annotations for category-level concept specification, while PACO offers annotations for fine-grained attribute-level concept specification. Each episode provides support-image bounding boxes as visual exemplars and requires the model to predict the target mask in the query image.


\subsection{Task Taxonomy}

As summarized in~\cref{tab:svps_tasks}, CASE encompasses six task families that evaluate different facets of positive-negative concept specification. The category-level tasks comprise {\em supercategory exclusion} (SuperEx), {\em fine-grained isolation within a supercategory} (FineIso), and {\em functional union across related categories} (Union). The attribute-level tasks include {\em exclusion} (AttrEx), {\em isolation} (AttrIso), and {\em cross-category union} (AttrUnion). Together, these tasks assess whether a model can move beyond naive positive matching and leverage negative exemplars as exclusion or boundary constraints for precise target concept specification.


\subsection{Episode Construction}

Following the standard FSS protocol, CASE is organized in an episodic format. Each episode consists of a support set with positive and negative bounding-box prompts, a query image, and a binary ground-truth query mask. The support prompts specify the target concept, and the query mask serves as the expected segmentation output in the query image for evaluation.


To construct an episode, we first instantiate a positive-negative concept definition using COCO or PACO annotations, such as a supercategory paired with an excluded class, a function-driven category group, or an attribute-defined object group. We then sample support images containing the required positive and negative exemplars. Query images are sampled independently to contain target objects and optionally include distractors. The target mask is generated from the underlying annotations according to the instantiated concept definition.


This design prevents the task from being solved by simply matching a positive support example. Because negative exemplars explicitly define what should be rejected or separated from the target, the model must infer the intended concept from the interplay between positive and negative prompts. Detailed per-task statistics and construction parameters are provided in the supplementary material, Sec.~\emph{CASE Benchmark Construction}.

\section{Method}
\label{sec:method}

\subsection{Task Formulation}


Specified Visual Prompt Segmentation (SVPS) extends conventional FSS by defining the target concept through both positive and negative visual exemplars. Specifically, each {\em episode} contains a $K$-shot support set $\mathcal{S}=\{(I_i^{s},\mathcal{B}_i^{+},\mathcal{B}_i^{-})\}_{i=1}^{K}$ and a query image $I^{q}$ with the ground-truth mask $M^{q}$. The positive bounding boxes $\mathcal{B}_i^{+}$ enclose visual evidence that must be included, whereas the negative boxes $\mathcal{B}_i^{-}$ indicate irrelevant distractors to be excluded. 
Given these inputs, the model $f_{\theta}$, parameterized by $\theta$, predicts a binary mask $\hat{M}^{q}$ for the query image:
\begin{equation}
\hat{M}^{q}
=
f_{\theta}(\mathcal{S}, I^{q}).
\end{equation}
Conventional FSS can be viewed as a special case of SVPS where $\mathcal{B}_i^{-}=\emptyset$ for all support examples. Therefore, an SVPS model should leverage negative exemplars to suppress distractors when they are provided, while retaining competitive segmentation performance under the positive-only setting.


\subsection{Model Overview}

\begin{figure*}[t]
\vspace{-2mm}
\centering
\includegraphics[width=0.9\linewidth]{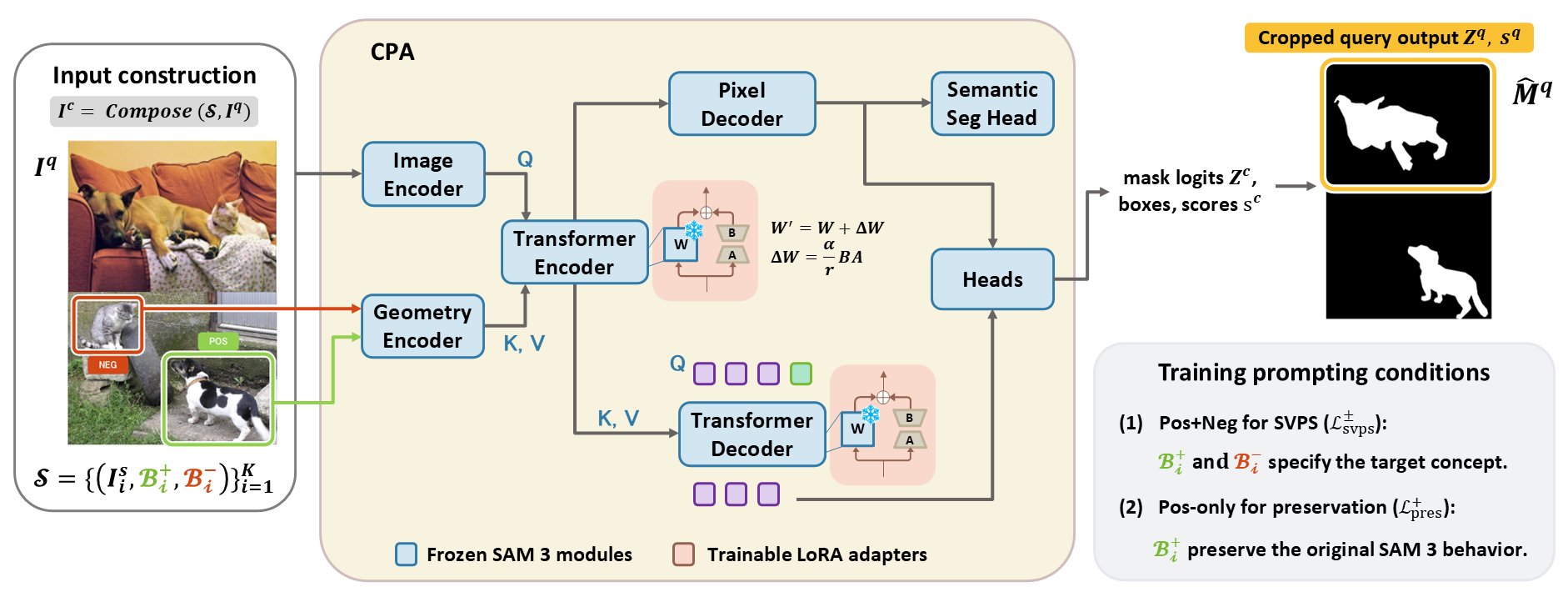}
\caption{
\textbf{Overview of Concept Prompt Adaptation (CPA).} 
A $K$-shot support set $\mathcal{S}$ and query image $I^{q}$ are composed into a shared canvas $I^{c}$, with positive and negative support boxes used as visual prompts.
CPA augments the original SAM 3 detector with two trainable LoRA modules while keeping the original SAM 3 weights frozen.
The final query prediction is obtained from the cropped query outputs $Z^{q}$ and $s^{q}$.
The inset summarizes the two prompting conditions used during training; their formal objectives are defined in \cref{sec:training_objective}.
}
\label{fig:model_overview}
\vspace{-4mm}
\end{figure*}

\cref{fig:model_overview} illustrates our \textbf{Concept Prompt Adaptation (CPA)} pipeline. CPA adapts SAM 3~\cite{carion2026sam3} to SVPS by augmenting the original SAM 3 detector with two trainable LoRA~\cite{hu2021lora} modules while keeping the original SAM 3 weights frozen. We separate the pipeline into two parts: (i) the shared-canvas input construction from the support set $\mathcal{S}$ and query image $I^{q}$, and (ii) the CPA-adapted SAM 3 forward path, where one LoRA module is attached to the transformer encoder and the other is attached to the transformer decoder.

Given $\mathcal{S}$ and $I^{q}$, CPA first composes them into a shared canvas $I^{c}$ and transforms the positive and negative boxes into the canvas coordinate system~\cite{tsai2026fsssam3}. The canvas image is processed by the SAM 3 image encoder, while the transformed prompt boxes are processed by the geometry encoder. The resulting features are passed through the SAM 3 transformer encoders and decoders, then through the prediction heads. The detector outputs mask logits $Z^{c}$, boxes, and confidence scores $s^{c}$ on the composite canvas. We crop the query region from $Z^{c}$ to obtain query logits $Z^{q}$, with corresponding confidence scores denoted as $s^{q}$. The final query prediction $\hat{M}^{q}$ is derived from $Z^{q}$.



During training, CPA is optimized under two prompting conditions. In the positive-negative condition, the adapted model receives both $\mathcal{B}^{+}$ and $\mathcal{B}^{-}$ and is supervised by the CASE target mask $M^{q}$. In the positive-only preservation condition, only $\mathcal{B}^{+}$ is provided, and the adapted model is regularized to remain close to the original frozen SAM 3 prediction under the same positive-only input. The following subsections describe the shared-canvas formulation, LoRA placement in CPA, and the training objective, respectively.



\subsection{Shared-Canvas Prompting}

Following FSS-SAM3~\cite{tsai2026fsssam3}, we adopt a shared-canvas formulation to process the support and query images within a unified spatial and attention space. The support set and query image are composed into a single image-like canvas:
\begin{equation}
I^{c}
=
\mathrm{Compose}(\mathcal{S}, I^{q}),
\end{equation}
and all support boxes $\{\mathcal{B}_i^{+},\mathcal{B}_i^{-}\}_{i=1}^{K}$ are transformed into the coordinate system of $I^{c}$. The composite image and box prompts are then fed into SAM 3.

This formulation allows support and query regions to interact within the same attention space while reusing SAM 3's native box-prompt interface, eliminating the need for an explicit module. The SAM 3 detector heads output composite mask logits $Z^{c}$, boxes, and confidence scores $s^{c}$. We crop the query region from the composite logits to obtain query logits $Z^{q}=\mathrm{Crop}_{q}(Z^{c})$, and use the corresponding confidence scores $s^{q}$ to select or rank predictions. The binary query mask is obtained from the selected query logits as $\hat{M}^{q}=\mathbf{1}[\sigma(Z^{q})>\tau]$, where $\sigma(\cdot)$ denotes the sigmoid function and $\tau$ is the mask threshold. For CASE episodes, both positive and negative boxes are used. For positive-only evaluation and preservation regularization, only positive boxes are provided.



\subsection{LoRA-Based Concept Prompt Adaptation}



Frozen SAM 3 already provides strong positive-only few-shot segmentation behavior, but it does not interpret negative visual prompts as concept-level exclusion cues. To address this limitation, CPA adapts SAM 3 to positive-negative prompting while mitigating degradation of its original positive-only behavior. Let $f$ denote the original frozen SAM 3. CPA constructs an adapted model $f_{\theta}$ by augmenting $f$ with two trainable LoRA modules: one attached to the transformer encoder and the other attached to the transformer decoder. All original SAM 3 parameters remain frozen, and $\theta$ contains only the parameters of these LoRA modules. For a pretrained attention projection matrix
$W \in \mathbb{R}^{d_{\mathrm{out}} \times d_{\mathrm{in}}}$, each LoRA module replaces the effective projection by
\begin{equation}
W'
=
W + \Delta W,
\quad
\Delta W
=
\frac{\alpha}{r}BA,
\end{equation}
where
$A \in \mathbb{R}^{r \times d_{\mathrm{in}}}$ and
$B \in \mathbb{R}^{d_{\mathrm{out}} \times r}$
are trainable matrices of rank $r \ll \min(d_{\mathrm{in}}, d_{\mathrm{out}})$, and
$\alpha$ is a scaling factor. The LoRA modules shown in \cref{fig:model_overview} correspond to low-rank updates injected into the transformer encoder and decoder.

\subsection{Training Objective}
\label{sec:training_objective}

CPA is optimized with two objectives: an SVPS segmentation loss for positive-negative prompting and a preservation loss for maintaining positive-only behavior. Let $f_{\theta}$ denote the adapted model and $f$ denote the original frozen SAM 3 model. Given positive and negative prompts, the adapted model predicts query logits $Z^{q}_{\theta,\pm}$. We supervise this prediction using the ground-truth query mask $M^{q}$:
\begin{equation}
\label{eq:loss_svps}
\mathcal{L}_{\mathrm{svps}}^{\pm}
=
\mathcal{L}_{\mathrm{mask}}(Z^{q}_{\theta,\pm}, M^{q}),
\mbox{ where } 
Z^{q}_{\theta,\pm} = \mathrm{Crop}_{q}(f_{\theta}(I^c, \mathcal{B}^{+}, \mathcal{B}^{-}))
\end{equation}
and $\mathcal{L}_{\mathrm{mask}}$ combines sigmoid focal loss and Dice loss. The superscript $\pm$ indicates that the prediction is conditioned on both positive and negative prompts.

The preservation loss is computed under the positive-only prompting condition. For the same episode, the original frozen model $f$ produces reference query logits $Z^{q}_{f,+}$ using only positive boxes. Under the same prompting condition, the adapted model $f_{\theta}$ outputs query logits $Z^{q}_{\theta,+}$. We penalize the discrepancy between the two probability maps using the squared $L_2$ norm:
\begin{gather}
\label{eq:loss_pres}
\mathcal{L}_{\mathrm{pres}}^{+}
=
\left\|
\sigma(Z^{q}_{\theta,+})
-
\sigma(Z^{q}_{f,+})
\right\|_2^2,
\mbox{ where}\\
Z^{q}_{\theta,+} = \mathrm{Crop}_{q}(f_{\theta}(I^c, \mathcal{B}^{+}))
\mbox{ and }
Z^{q}_{f,+} = \mathrm{Crop}_{q}(f(I^c, \mathcal{B}^{+})),
\end{gather}
and $\sigma(\cdot)$ denotes the sigmoid function. The superscript $+$ indicates that this term is computed under the positive-only prompting condition.

The final training objective is:
\begin{equation}
\label{eq:loss_total}
\mathcal{L}
=
\mathcal{L}_{\mathrm{svps}}^{\pm}
+
\lambda_{\mathrm{pres}}
\mathcal{L}_{\mathrm{pres}}^{+},
\end{equation}
where $\lambda_{\mathrm{pres}}$ controls the strength of positive-only preservation.

\section{Experiments}
\label{sec:experiments}


\subsection{Experimental Setup}

\subsubsection{Datasets.}
We evaluate CPA under two settings.

\paragraph{CASE.}
CASE evaluates specified visual prompt segmentation with positive and negative visual exemplars. It contains six task families constructed from COCO~\cite{lin2014coco} and PACO~\cite{ramanathan2023paco}: SuperEx, FineIso, and Union for category-level concepts, and AttrEx, AttrIso, and AttrUnion for attribute-level concepts. Each episode provides positive and negative support boxes, a query image, and a binary target mask defined by a structured concept specification, such as category exclusion, fine-grained isolation, functional union, attribute isolation, attribute union, or attribute exclusion. The evaluation split is fixed at the episode level to ensure deterministic comparison across models.

\paragraph{Standard FSS benchmark.}
We evaluate the CASE-trained CPA checkpoints directly on all four COCO-$20^i$ folds~\cite{Nguyen_2019_ICCV,lin2014coco}, without retraining or fold-specific tuning. COCO-$20^i$ is a standard few-shot segmentation benchmark derived from COCO, where the 80 object categories are divided into four folds with 20 categories per fold. We report results under both 1-shot and 5-shot settings. Since standard FSS uses only positive support annotations, each support mask is converted into its tight bounding box and used as a positive visual prompt.

\subsubsection{Evaluation Metrics.}
We use mean Intersection-over-Union (mIoU) as the primary metric. For CASE, we report task-wise mIoU and the average mIoU across all tasks. We further aggregate results into category-level and attribute-level groups. For COCO-$20^i$, we report fold-wise and mean mIoU under the standard positive-only 1-shot and 5-shot protocols.

\subsubsection{Implementation Details.}
CPA is built on SAM 3 using the shared-canvas formulation of FSS-SAM3~\cite{tsai2026fsssam3}, where support and query images are composed into a single $1008 \times 1008$ canvas. Positive and negative visual prompts are represented as support bounding boxes. We keep the pretrained SAM 3 weights frozen and insert LoRA modules into transformer encoder and decoder layers, resulting in 1.13M trainable parameters. CPA is trained with AdamW at a learning rate of $1 \times 10^{-4}$, with LoRA rank $r=16$ and scaling factor $\alpha=32$. Standard FSS evaluation follows the positive-only prompting protocol, which is also used for the preservation regularization. Query masks are binarized at $\tau=0.5$. Additional details are provided in the supplementary material, Sec.~\emph{Implementation Details}.

\subsection{Main Results on CASE}

\begin{table*}[t]
\vspace{-2mm}
\centering
\caption{\textbf{CASE benchmark results:} task-wise and average mIoU across six tasks.}
\label{tab:case_main}
\scriptsize
\begin{tabular*}{\textwidth}{@{\extracolsep{\fill}} l c c c c c c c c @{}}
\toprule
Method & Prompt & SuperEx & FineIso & Union & AttrEx & AttrIso & AttrUnion & \textbf{Mean} \\
\midrule
\multicolumn{9}{c}{\textbf{1-shot}} \\
\midrule
GF-SAM~\cite{zhang2024gfsam} & Pos Only & 24.87 & 35.61 & 33.25 & 16.74 & 17.44 & 10.26 & 23.03 \\
FSSAM~\cite{xu2025fssam} & Pos Only & 28.55 & 38.97 & 35.89 & 17.28 & 15.87 & 14.83 & 25.23 \\
SANSA~\cite{cuttano2025sansa} & Pos Only & 26.96 & 53.65 & 29.83 & 15.65 & 22.48 & 17.07 & 27.61 \\
FSS-SAM3~\cite{tsai2026fsssam3} & Pos Only & 28.80 & 56.23 & 43.43 & 20.05 & 19.83 & 15.32 & 30.61 \\
FSS-SAM3~\cite{tsai2026fsssam3} & Pos+Neg & 26.22 & 20.43 & 30.52 & 17.06 & 6.33 & 11.17 & 18.62 \\
\midrule
CPA (Ours) & Pos Only & 27.04 & 46.64 & 38.11 & 20.29 & 24.46 & 14.36 & 28.48 \\
\textbf{CPA (Ours)} & \textbf{Pos+Neg} & \textbf{37.11} & \textbf{56.70} & \textbf{52.07} & \textbf{21.98} & \textbf{28.74} & \textbf{18.39} & \textbf{35.83} \\
\midrule
\multicolumn{9}{c}{\textbf{5-shot}} \\
\midrule
GF-SAM~\cite{zhang2024gfsam} & Pos Only & 35.60 & 49.44 & 45.36 & 20.87 & 22.76 & 19.91 & 32.32 \\
FSSAM~\cite{xu2025fssam} & Pos Only & 37.11 & 55.13 & 44.51 & 17.76 & 17.87 & 20.61 & 32.17 \\
SANSA~\cite{cuttano2025sansa} & Pos Only & 34.37 & 61.55 & 36.89 & 18.45 & 25.24 & 23.87 & 33.40 \\
FSS-SAM3~\cite{tsai2026fsssam3} & Pos Only & 39.07 & 62.38 & 53.72 & 23.11 & 25.21 & 25.13 & 38.10 \\
FSS-SAM3~\cite{tsai2026fsssam3} & Pos+Neg & 39.76 & 50.83 & 47.50 & 20.74 & 21.48 & 24.55 & 34.14 \\
\midrule
CPA (Ours) & Pos Only & 45.85 & 62.23 & 49.23 & 25.82 & 25.83 & 26.84 & 39.30 \\
\textbf{CPA (Ours)} & \textbf{Pos+Neg} & \textbf{49.39} & \textbf{66.00} & \textbf{56.05} & \textbf{26.08} & \textbf{27.64} & \textbf{27.71} & \textbf{42.15} \\
\bottomrule
\end{tabular*}
\vspace{-2mm}
\end{table*}

\cref{tab:case_main} shows that CASE exposes limitations that are not visible in standard positive-only FSS. Positive-only methods remain competitive on tasks close to category matching, especially FineIso, but they cannot explicitly encode what should be rejected. This becomes clear on exclusion, union, and attribute-based tasks, where a target concept cannot be specified by positive similarity alone. Directly adding negative prompts to frozen SAM 3 is also ineffective: FSS-SAM3 drops from 30.61 to 18.62 in 1-shot and from 38.10 to 34.14 in 5-shot. This suggests that the frozen model is not calibrated to treat negative boxes as concept-level exclusion cues; instead, they can act as unstable suppression signals.

By adapting SAM 3 to use negative exemplars relative to positive anchors, CPA improves over its positive-only variant from 28.48 to 35.83 in 1-shot and from 39.30 to 42.15 in 5-shot. The improvement is consistent across both category-level and attribute-level tasks, showing that lightweight adaptation enables the model to interpret positive and negative supports as a structured concept specification rather than as independent box prompts.

\begin{table*}[t]
\centering
\caption{\textbf{Aggregated CASE results} by category-level and attribute-level concepts.}
\label{tab:case_summary}
\scriptsize
\begin{tabular*}{\textwidth}{@{\extracolsep{\fill}} ll ccc ccc @{}}
\toprule
\multirow{2}{*}{Method} & \multirow{2}{*}{Prompt} & \multicolumn{3}{c}{1-shot} & \multicolumn{3}{c}{5-shot} \\
\cmidrule(r){3-5} \cmidrule(l){6-8}
& & Category & Attribute & \textbf{Mean} & Category & Attribute & \textbf{Mean} \\
\midrule
GF-SAM~\cite{zhang2024gfsam} & Pos Only & 31.24 & 14.81 & 23.03 & 43.47 & 21.18 & 32.32 \\
FSSAM~\cite{xu2025fssam} & Pos Only & 34.47 & 15.99 & 25.23 & 45.58 & 18.75 & 32.17 \\
SANSA~\cite{cuttano2025sansa} & Pos Only & 36.81 & 18.40 & 27.61 & 44.27 & 22.52 & 33.40 \\
FSS-SAM3~\cite{tsai2026fsssam3} & Pos Only & 42.82 & 18.40 & 30.61 & 51.72 & 24.48 & 38.10 \\
FSS-SAM3~\cite{tsai2026fsssam3} & Pos+Neg & 25.72 & 11.52 & 18.62 & 46.03 & 22.26 & 34.14 \\
\midrule
CPA (Ours) & Pos Only & 37.26 & 19.70 & 28.48 & 52.44 & 26.16 & 39.30 \\
\textbf{CPA (Ours)} & \textbf{Pos+Neg} & \textbf{48.63} & \textbf{23.04} & \textbf{35.83} & \textbf{57.15} & \textbf{27.14} & \textbf{42.15} \\
\bottomrule
\end{tabular*}
\vspace{-4mm}
\end{table*}

As shown in \cref{tab:case_summary}, foundation models favor category-level over attribute-level concepts due to object-centric priors. This highlights the value of CASE as a broader benchmark evaluating visual exclusions, compositional groupings, and attribute-defined targets. Comparing CPA with and without negative prompts shows that negative exemplars provide complementary support, improving the CASE mean from 28.48 to 35.83 in 1-shot and from 39.30 to 42.15 in 5-shot. The gains are more pronounced on category-level concepts, suggesting that negative prompts are especially useful when the target set can be refined by excluding related object categories or disambiguating broad concept groups. Attribute-level gains are smaller but consistent, indicating that material and color reasoning remains difficult. Detailed breakdowns are provided in the supplementary material, Sec.~\emph{Additional Experimental and Qualitative Results}.

\subsection{Positive-Only FSS Evaluation}

\begin{table*}[t]
\vspace{-2mm}
  \centering
  \caption{\textbf{Comparison with state-of-the-art FSS methods on COCO-$20^{i}$.} CPA is evaluated under the standard positive-only protocol.}
  \label{tab:results_coco}
  \scriptsize
  \begin{tabular*}{\textwidth}{@{\extracolsep{\fill}} ll ccccc ccccc @{}}
    \toprule
    \multirow{2}{*}{Method} & \multirow{2}{*}{Type} & \multicolumn{5}{c}{1-shot} & \multicolumn{5}{c}{5-shot} \\
    \cmidrule(r){3-7} \cmidrule(l){8-12}
    & & $20^0$ & $20^1$ & $20^2$ & $20^3$ & Mean & $20^0$ & $20^1$ & $20^2$ & $20^3$ & Mean \\
    \midrule
    PFENet~\cite{tian2020pfenet} & Classical FSS & 36.5 & 38.6 & 34.5 & 33.8 & 35.8 & 36.5 & 43.3 & 37.8 & 38.4 & 39.0 \\
    HSNet~\cite{min2021hsnet} & Classical FSS & 36.3 & 43.1 & 38.7 & 38.7 & 39.2 & 43.3 & 51.3 & 48.2 & 45.0 & 46.9 \\
    BAM~\cite{lang2022bam} & Classical FSS & 43.4 & 50.6 & 47.5 & 43.4 & 46.2 & 49.3 & 54.2 & 51.6 & 49.6 & 51.2 \\
    HDMNet~\cite{peng2023hdmnet} & Classical FSS & 43.8 & 55.3 & 51.6 & 49.4 & 50.0 & 50.6 & 61.6 & 55.7 & 56.0 & 56.0 \\
    \midrule
    Matcher~\cite{liu2024matcher} & Foundation-based & 52.7 & 53.5 & 52.6 & 52.1 & 52.7 & 60.1 & 62.7 & 60.9 & 59.2 & 60.7 \\
    GF-SAM~\cite{zhang2024gfsam} & Foundation-based & 56.6 & 61.4 & 59.6 & 57.1 & 58.7 & 67.1 & 69.4 & 66.0 & 64.8 & 66.8 \\
    SANSA~\cite{cuttano2025sansa} & Foundation-based & 58.9 & 62.6 & 61.5 & 58.0 & 60.2 & -- & -- & -- & -- & 67.3 \\
    FSSAM~\cite{xu2025fssam} & Foundation-based & 59.9 & 65.6 & 62.1 & 61.6 & 62.3 & 68.6 & 74.0 & 64.5 & 69.9 & 69.3 \\
    FSS-SAM3~\cite{tsai2026fsssam3} & Foundation-based & \textbf{64.2} & \textbf{66.8} & \textbf{67.7} & \textbf{67.5} & \textbf{66.6} & 68.6 & 72.5 & 72.5 & 73.6 & 71.8 \\
    \midrule
    \textbf{CPA (Ours)} & Ours & 61.1 & 64.3 & 59.2 & 61.6 & 61.6 & \textbf{74.4} & \textbf{74.1} & \textbf{74.6} & \textbf{76.2} & \textbf{74.8} \\
    \bottomrule
  \end{tabular*}
  \vspace{-2mm}
\end{table*}

Although CPA is trained for SVPS, we evaluate its positive-only FSS behavior on COCO-$20^i$ under the standard protocol, where support masks are converted into tight bounding boxes. As shown in \cref{tab:results_coco}, CPA obtains 61.6 mIoU in 1-shot, which is lower than FSS-SAM3. However, CPA achieves the best 5-shot performance, 74.8 mIoU, among all methods. This suggests that the adapted model remains responsive to positive visual prompts and can effectively exploit additional support evidence when multiple exemplars are available.

\subsection{Ablation and Sensitivity Analysis}

\begin{table}[t]
\centering
\caption{\textbf{Effect of positive-only preservation.} CASE is evaluated with positive-negative prompts, while COCO-$20^i$ is evaluated under the standard FSS protocol.}
\label{tab:ablation_pres}
\scriptsize
\setlength{\tabcolsep}{6pt}
\begin{tabular}{@{} l c c c c c @{}}
\toprule
\multirow{2}{*}{Method} & \multicolumn{3}{c}{CASE} & \multicolumn{2}{c}{COCO-$20^i$} \\
\cmidrule(r){2-4} \cmidrule(l){5-6}
& Category & Attribute & Mean & 1-shot & 5-shot \\
\midrule
FSS-SAM3~\cite{tsai2026fsssam3} & 25.7 & 11.5 & 18.6 & \textbf{66.6} & 71.8 \\
CPA w/o Preservation & \textbf{50.6} & \textbf{25.2} & \textbf{37.9} & 50.8 & 66.8 \\
CPA & 48.6 & 23.0 & 35.8 & 61.6 & \textbf{74.8} \\
\bottomrule
\end{tabular}
\vspace{-4mm}
\end{table}

\subsubsection{Effect of preservation regularization.}
\cref{tab:ablation_pres} shows the adaptation--preservation trade-off. Removing preservation gives the best CASE performance, but substantially degrades COCO-$20^i$ positive-only FSS. Adding preservation slightly lowers CASE mIoU, yet recovers much of the positive-only FSS ability and yields strong 5-shot COCO performance. We therefore use the preservation-regularized CPA as the default, since it improves SVPS while maintaining the usability of SAM 3 under conventional positive prompts.

\begin{table}[t]
\vspace{-2mm}
\centering
\caption{\textbf{Negative prompt sensitivity on CASE-NegStress.} We report mIoU (\%). The full per-task results are provided in the supplementary material.}
\label{tab:neg_prompt_sensitivity}
\scriptsize
\setlength{\tabcolsep}{5pt}
\begin{tabular}{@{} l l c c c c @{}}
\toprule
Group & Method & 0 Neg & 1 Neg & 2 Neg & All Neg \\
\midrule
\multirow{2}{*}{Category}
& FSS-SAM3~\cite{tsai2026fsssam3} & 37.28 & 29.82 & 21.86 & 12.53 \\
& CPA & \textbf{49.29} & \textbf{59.66} & \textbf{60.27} & \textbf{59.19} \\
\midrule
\multirow{2}{*}{Attribute}
& FSS-SAM3~\cite{tsai2026fsssam3} & 10.04 & 8.68 & 5.40 & 2.74 \\
& CPA & \textbf{18.22} & \textbf{24.12} & \textbf{25.49} & \textbf{22.75} \\
\midrule
\multirow{2}{*}{Overall}
& FSS-SAM3~\cite{tsai2026fsssam3} & 23.66 & 19.25 & 13.63 & 7.63 \\
& CPA & \textbf{33.76} & \textbf{41.89} & \textbf{42.88} & \textbf{40.97} \\
\bottomrule
\end{tabular}
\vspace{-4mm}
\end{table}

\subsubsection{Negative prompt sensitivity.}
CASE-NegStress is a dedicated 1-shot subset that fixes the query image, target mask, positive prompts, and candidate negative pool while varying only the negative-prompt count; 0-Neg denotes Pos-only on this subset, not full CASE. As shown in \cref{tab:neg_prompt_sensitivity}, FSS-SAM3 degrades sharply as more negatives are added, dropping from 23.66 to 7.63 overall mIoU. In contrast, CPA improves from 33.76 to 40.97. This indicates that CPA learns to interpret negative prompts relative to positive anchors, while the original SAM 3 treats them as unstable suppression signals.

\subsection{Qualitative Results and Limitations}

\begin{figure}[t]
\vspace{-2mm}
\centering
\includegraphics[width=0.8\linewidth]{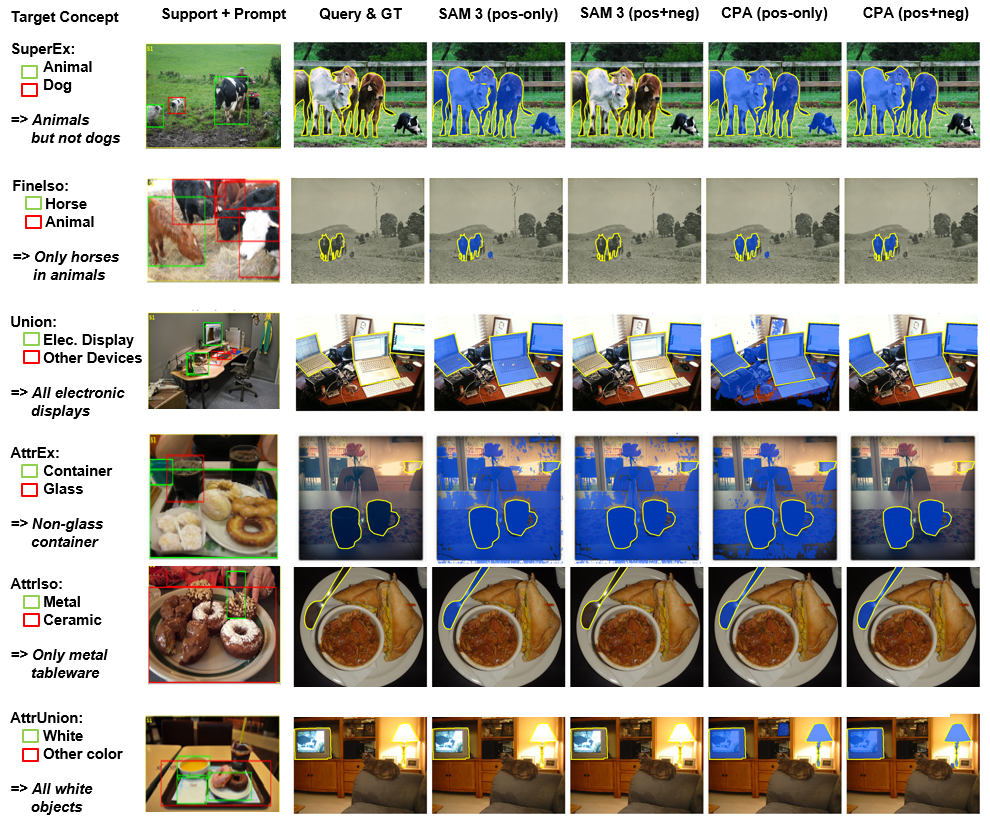}
\caption{
\textbf{Qualitative comparison on CASE.} Green/red boxes in the support image indicate positive/negative visual prompts. Yellow contours denote ground-truth query masks, and blue overlays denote model predictions. We compare SAM 3 (FSS-SAM3) and CPA under positive-only and positive-negative prompting.
}
\label{fig:qualitative}
\vspace{-4mm}
\end{figure}

\subsubsection{Qualitative Results.}
\cref{fig:qualitative} presents representative qualitative comparisons on CASE. SAM 3 with positive-only prompting often misses valid target regions or fails to respect the intended exclusion or attribute constraint. Naively adding negative prompts to SAM 3 remains unstable and may suppress valid target regions together with distractors. CPA with positive-only prompting remains responsive to positive prompts, but may retain distractor regions because the target concept remains underspecified without negative exemplars, especially for exclusion and attribute-based tasks. CPA with positive-negative prompting provides the most consistent predictions, producing masks that better match the concept jointly specified by positive and negative visual exemplars.

\subsubsection{Limitations.}
Attribute-level tasks remain challenging when color or material cues are subtle or ambiguously annotated, and cross-category attribute union is particularly difficult because the model must group visually diverse objects by a shared property rather than by category or shape. CASE is also limited by the categories, masks, and attributes available in COCO and PACO. Additional qualitative results and failure cases are provided in the supplementary material, Secs.~\emph{Additional Qualitative Results} and \emph{Failure Cases and Limitations}.

\section{Conclusion}

We introduced \textbf{Specified Visual Prompt Segmentation (SVPS)}, a new segmentation setting, where positive and negative visual exemplars jointly specify target concepts. To evaluate this setting, we constructed \textbf{CASE} (\textbf{C}oncept-\textbf{A}ware \textbf{S}egmentation from Positive and Negative Visual \textbf{E}xemplars), a benchmark built upon COCO and PACO annotations with six category- and attribute-level task families. 
Our experiments show that while frozen SAM 3 exhibits strong few-shot segmentation capabilities with solely positive prompts, it struggles to reliably interpret negative visual exemplars as exclusion evidence. To address this limitation, we presented \textbf{Concept Prompt Adaptation (CPA)}, a lightweight LoRA-based adaptation baseline to enhance positive-negative prompt interpretation. CPA significantly improves performance on CASE while keeping competitive performance under positive-only FSS setups. We believe that CASE can serve as a useful testbed for future research on positive-negative visual prompting, negative exemplars, and concept-level visual prompt segmentation.


\vspace{-1mm}


\subsubsection{\textbf{Acknowledgements.}}This work was supported in part by the National Science and Technology Council (NSTC) under grants 114-2221-E-A49-038-MY3, 115-2221-E-A49-119-MY3, 115-2628-E-001-002 and 115-2634-F-A49-011, by IIS grant 3006-C4741, and by MediaTek.

%
%
\bibliographystyle{splncs04}
\bibliography{main}
\setcounter{page}{1}       
\setcounter{figure}{0}     
\setcounter{table}{0}      
\renewcommand{\thepage}{S\arabic{page}}     
\renewcommand{\thefigure}{S\arabic{figure}} 
\renewcommand{\thetable}{S\arabic{table}}   


\title{Supplementary Material of CASE: Concept-Aware Segmentation from Positive and Negative Visual Exemplars} 

\author{Yi-Jen Tsai\inst{1} \and
Yen-Yu Lin\inst{1} \and
Chien-Yao Wang\inst{2}\thanks{Corresponding author}}
\authorrunning{Y.-J. Tsai et al.}
\institute{National Yang Ming Chiao Tung University, Taiwan\\
\email{tsai.cs14@nycu.edu.tw, lin@cs.nycu.edu.tw} \and
Institute of Information Science, Academia Sinica, Taiwan\\
\email{kinyiu@iis.sinica.edu.tw}}

\maketitle

\section{CASE Benchmark Construction}

This section provides additional details on the construction of CASE. CASE is designed to evaluate whether a segmentation model can infer a target concept from both positive and negative visual exemplars. Unlike standard few-shot semantic segmentation benchmarks, where the support set usually defines a single positive class, CASE episodes require the model to parse compositional intent relations such as inclusion, exclusion, fine-grained isolation, functional grouping, attribute selection, attribute union, and attribute exclusion.

\subsection{Data Sources}

CASE is built from two annotation sources. The category-level tasks, including SuperEx, FineIso, and Union, are constructed from MSCOCO instance segmentation annotations~\cite{lin2014coco}. We use COCO train2017 to generate candidate training episodes and COCO val2017 to generate fixed evaluation episodes. The attribute-level tasks, including AttrIso, AttrUnion, and AttrEx, are constructed from PACO-LVIS annotations~\cite{ramanathan2023paco}. PACO provides object-level category annotations and attribute labels, which allow us to define visual concepts based on color and material properties.

For PACO-based tasks, we use object-level categories and filter out part-level categories whose category names contain a part delimiter. This avoids constructing episodes in which the target concept is defined over object parts rather than complete object instances. We further restrict attribute concepts to visually grounded color and material attributes. The color set contains black, blue, brown, green, grey, orange, pink, purple, red, white, and yellow. The material set contains stone, wood, rattan, fabric, crochet, wool, leather, velvet, metal, paper, plastic, glass, and ceramic.

To specify, not every COCO/PACO image is used: only images satisfying the task-specific support and target constraints are eligible. Episode and mask construction is programmatic from the source annotations; only the high-level Union group definitions are manually specified.

\begin{table}[t]
\centering
\caption{\textbf{Data sources used for each CASE task family.} Category-level tasks are constructed from COCO, while attribute-level tasks are constructed from PACO-LVIS.}
\label{tab:supp_data_sources}
\scriptsize
\begin{tabular}{@{}lll@{}}
\toprule
Task family & Source annotation & Split usage \\
\midrule
SuperEx   & COCO instance segmentation & train2017 / val2017 \\
FineIso   & COCO instance segmentation & train2017 / val2017 \\
Union     & COCO instance segmentation & train2017 / val2017 \\
AttrIso   & PACO-LVIS object attributes & train / val \\
AttrUnion & PACO-LVIS object attributes & train / val \\
AttrEx    & PACO-LVIS object attributes & train / val \\
\bottomrule
\end{tabular}
\end{table}

\subsection{Episode Format}

Each CASE episode in the benchmark contains a target image, a frozen support set, a task label, an intent description, a frozen target mask, and frozen positive/negative visual prompts. The format is designed for reproducible evaluation: the support examples, prompt boxes, and target masks are fixed in advance, so different methods are evaluated on exactly the same episodes.

A support image contains positive and negative bounding-box prompts. Positive prompts indicate visual exemplars that should be included in the target concept, while negative prompts indicate exemplars that should be excluded or used as discriminative counterexamples. The target image contains a ground-truth mask encoded in COCO-style run-length encoding (RLE). This avoids requiring users to re-run the benchmark construction code.

A simplified released episode format is shown below.

\begin{verbatim}
{
"episode_id": "...",
"task": "SuperEx",
"intent": "All vehicle MINUS car",
"target": {
"image_id": ...,
"image_path": "...",
"gt_foreground_pixels": ...,
"gt_mask_rle": {
"size": [height, width],
"counts": "..."
},
"pos_bboxes": [...],
"neg_bboxes": [...]
},
"support": [
{
"image_id": ...,
"image_path": "...",
"prompts": {
"pos_bboxes": [...],
"neg_bboxes": [...]
}
}
]
}
\end{verbatim}

In the internal construction stage, each episode is first generated with a task-specific dynamic ground-truth configuration. This configuration specifies how the target mask should be produced from the underlying COCO or PACO annotations. For example, SuperEx stores a base supercategory and an excluded category, FineIso stores a target category, Union stores a fixed set of positive categories, AttrIso stores a target object category and a target attribute, AttrUnion stores a target attribute, and AttrEx stores a semantic object group and an excluded attribute. During release compilation, this dynamic configuration is converted into a frozen RLE target mask.

\subsection{Benchmark Statistics}

\begin{table}[t]
\centering
\caption{\textbf{Detailed statistics of the CASE benchmark}. We report the number of candidate training episodes, fixed evaluation episodes, concept definitions, and average prompts per episode for each task family. The overall average prompt counts are weighted by the number of candidate training episodes.}
\label{tab:supp_case_stats}
\scriptsize
\begin{tabular}{@{} l c c c c c @{}}
\toprule
Task & Source & \#Train Ep. & \#Eval Ep. & \#Concepts & Avg. Pos / Neg \\
\midrule
SuperEx   & COCO      & 156,000 & 9,004 & 78  & 6.71 / 2.31 \\
FineIso   & COCO      & 140,727 & 8,662 & 79  & 2.43 / 6.17 \\
Union     & COCO      &  50,000 & 1,806 & 25  & 5.04 / 3.96 \\
AttrEx    & PACO      &  12,909 & 1,528 & 109 & 2.09 / 1.04 \\
AttrIso   & PACO      &  66,942 & 1,100 & 94  & 1.04 / 2.09 \\
AttrUnion & PACO      &  34,812 & 1,116 & 21  & 2.15 / 1.64 \\
\midrule
\textbf{Overall} & COCO/PACO & \textbf{461,390} & \textbf{23,216} & \textbf{406} & \textbf{3.94 / 3.56} \\
\bottomrule
\end{tabular}
\end{table}

\cref{tab:supp_case_stats} provides detailed statistics for CASE. The training column reports candidate episodes used to sample training instances, while the evaluation column reports fixed episodes used for deterministic comparison. The number of concepts counts the distinct structured concept definitions within each task. The average prompt counts are computed across training episodes, and the overall average is weighted by the number of candidate training episodes per task.

\subsection{Train/Test Splits}

CASE follows the original train/validation split of its source annotations. COCO-based tasks use train2017 for candidate training episodes and val2017 for evaluation episodes. PACO-based tasks follow the PACO train/val split. Thus, the training and evaluation episodes are image-disjoint by construction. The split is not concept-disjoint: the same structured concept definition may appear in both splits, while the underlying images remain disjoint.

The evaluation split is frozen for reproducibility. For each validation episode, we first sample a 5-shot support set using a global random seed. The corresponding 1-shot episode is obtained by taking the first support image from the 5-shot set. Therefore, the 1-shot support set is nested within the 5-shot support set. This design ensures that the 1-shot and 5-shot protocols evaluate the same target episodes while differing only in the number of available support exemplars.

At training time, support images are dynamically sampled from a larger candidate support pool, which increases prompt diversity while keeping the underlying intent definition fixed. Unless otherwise specified, candidate training episodes are generated with a fixed random seed, and each concept is capped by a maximum number of target images to avoid severe imbalance.

\begin{table}[t]
\centering
\caption{\textbf{General construction parameters used in CASE.} The overall average prompt counts reported in the main paper are weighted by the number of candidate training episodes.}
\label{tab:supp_construction_params}
\scriptsize
\begin{tabular}{@{}ll@{}}
\toprule
Parameter & Value \\
\midrule
Minimum target foreground area & 500 pixels \\
Training support candidate pool size & 300 \\
Maximum training support pool size & 100 \\
Maximum target images per concept & 2000 \\
Training episode seed & 42 \\
Evaluation support sampling seed & 42 \\
Evaluation protocols & 1-shot and 5-shot \\
Support nesting & 1-shot $\subset$ 5-shot \\
Maximum positive prompts per support image & 3 \\
Maximum negative prompts per support image & 3 \\
\bottomrule
\end{tabular}
\end{table}

\subsection{Prompt Sampling Protocol}

CASE uses bounding boxes as visual prompts. Each support image may contain multiple positive and negative boxes. Positive boxes are selected from instances that satisfy the intended concept, whereas negative boxes are selected from semantically or visually related distractors. Thus, negatives are task-related distractors rather than arbitrary classes. All prompts correspond to object-level instances; CASE does not cover stuff categories. This design forces the model to infer not only what should be segmented, but also what should be rejected.

To reduce annotation noise and duplicated prompts, candidate boxes are filtered by an IoU-based duplicate suppression rule. A newly selected box is discarded if it overlaps too strongly with an existing prompt box. We also avoid selecting positive boxes that collide with negative boxes, since such overlaps would make the support intent ambiguous.

Training and evaluation differ in strictness. For training, we adopt a target-relaxed but support-strict protocol. A training target image is required to contain at least one positive foreground instance, but it is not always required to contain a negative distractor. This increases concept coverage and avoids over-restricting the training distribution. In contrast, the support set remains strict: support images are required to contain explicit positive and negative exemplars so that each training episode provides a clear contrastive intent.

For evaluation, we adopt a stricter discriminative protocol. Whenever the task definition involves exclusion or disambiguation, target images are required to contain both foreground objects satisfying the intent and related distractors. Support images are also required to contain positive and negative exemplars. As a result, the evaluation split directly tests whether a model can use negative visual prompts to suppress distractors rather than merely segmenting all objects that resemble the positive exemplars.

\subsection{Dynamic Ground-truth Mask Generation}

The target mask of each episode is generated from the source annotations according to its task-specific intent. For COCO-based tasks, masks are generated from instance segmentation annotations. For PACO-based tasks, masks are generated by filtering object annotations according to semantic object groups and visual attributes. The dynamic mask rules are summarized in Table~\ref{tab:supp_task_rules}. In exclusion tasks, the target mask is computed by subtracting the excluded instances from the base foreground. In union tasks, the target mask is the union of all instances that satisfy the positive concept. In attribute tasks, the target mask is obtained by selecting or excluding instances according to PACO attribute labels.

\begin{table*}[t]
\centering
\caption{\textbf{Task-specific construction rules in CASE.} Pos. and Neg. indicate the positive and negative support prompts, respectively.}
\label{tab:supp_task_rules}
\scriptsize
\begin{tabularx}{\textwidth}{@{}lXXXX@{}}
\toprule
Task & Intent & Target mask & Pos. prompts & Neg. prompts \\
\midrule
SuperEx
& All objects in a COCO supercategory except one excluded category.
& Union of all instances in the supercategory minus instances of the excluded category.
& Instances from non-excluded categories in the same supercategory.
& Instances from the excluded category. \\

FineIso
& Only one fine-grained category within a COCO supercategory.
& Instances of the target category.
& Instances of the target category.
& Instances from other categories in the same supercategory. \\

Union
& A fixed high-level category group, such as public transportation or rest furniture.
& Union of all instances belonging to the positive categories of the group.
& Instances from at least two positive categories when possible.
& Instances from fixed distractor categories associated with the group. \\

AttrIso
& Objects in a fixed object group with a specified color or material attribute.
& Instances in the target object group that possess the target attribute.
& Same-group instances with the target attribute.
& Same-group instances with different attributes. \\

AttrUnion
& All object instances with a specified color or material attribute, regardless of category.
& All valid object-level instances that possess the target attribute.
& Instances from multiple categories sharing the target attribute.
& Instances with different attributes from the same attribute family. \\

AttrEx
& Objects in a semantic object group excluding a specified color or material attribute.
& Instances in the semantic group minus instances with the excluded attribute.
& Instances with non-excluded attributes from the same attribute family.
& Instances with the excluded attribute. \\
\bottomrule
\end{tabularx}
\end{table*}

\section{Task-specific Details}

This section describes the six CASE task families in detail. Each task is designed to evaluate a different form of concept reasoning from positive and negative visual exemplars. In all task illustrations, green boxes on support images denote positive visual prompts and red boxes denote negative visual prompts. On target images, the green overlay denotes the ground-truth target mask. Red boxes on target images are shown only to indicate distractor instances for visualization and are not provided as model input during evaluation.

\subsection{SuperEx: Supercategory Exclusion}

\begin{figure}[t]
\centering
\includegraphics[width=\linewidth]{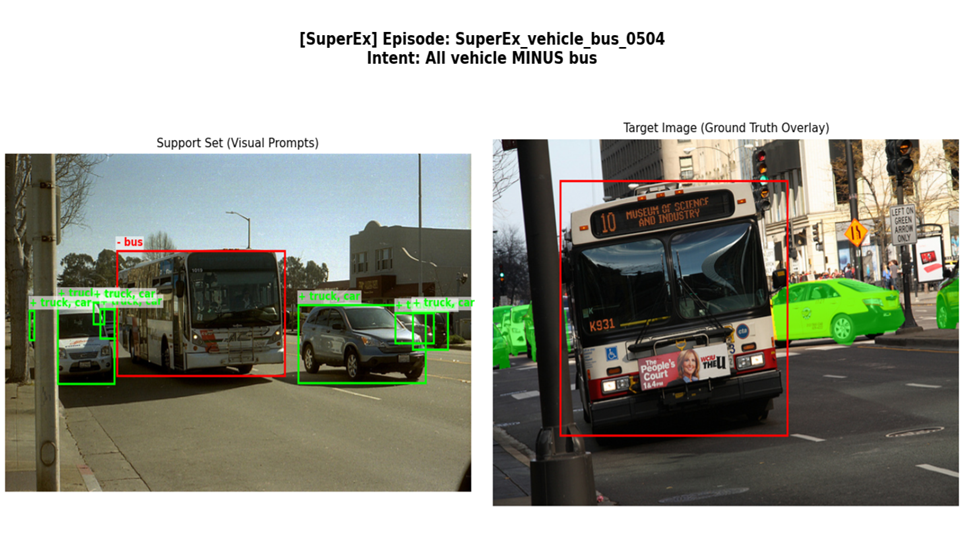}
\caption{
\textbf{Example SuperEx episode.} The intent is to segment all vehicles except buses.
}
\label{fig:superex}
\end{figure}

SuperEx evaluates whether a model can infer a broad category-level concept while excluding a specified fine-grained category. Each concept is defined as
\[
\mathcal{Y} = \mathcal{C}_{\mathrm{super}} \setminus \mathcal{C}_{\mathrm{neg}},
\]
where $\mathcal{C}_{\mathrm{super}}$ is a COCO supercategory and $\mathcal{C}_{\mathrm{neg}}$ is one excluded category within that supercategory. For example, an episode may define the intent ``all vehicles except cars.'' Positive prompts are sampled from non-excluded categories in the same supercategory, such as buses and trains, while negative prompts are sampled from the excluded category, such as cars.

The target mask is generated by first taking the union of all instances in the base supercategory and then subtracting all instances belonging to the excluded category:
\[
M = M(\mathcal{C}_{\mathrm{super}}) \setminus M(\mathcal{C}_{\mathrm{neg}}).
\]
This task tests whether a model can use negative exemplars to prevent over-generalization. A model that relies only on positive prompts may infer the broad supercategory and incorrectly include the excluded category.

For evaluation, target images are required to contain both valid foreground instances and excluded-category distractors. Support images are required to include at least two positive categories and at least one negative exemplar. This makes SuperEx a direct test of category-level exclusion.

\subsection{FineIso: Fine-grained Isolation}

\begin{figure}[t]
\centering
\includegraphics[width=\linewidth]{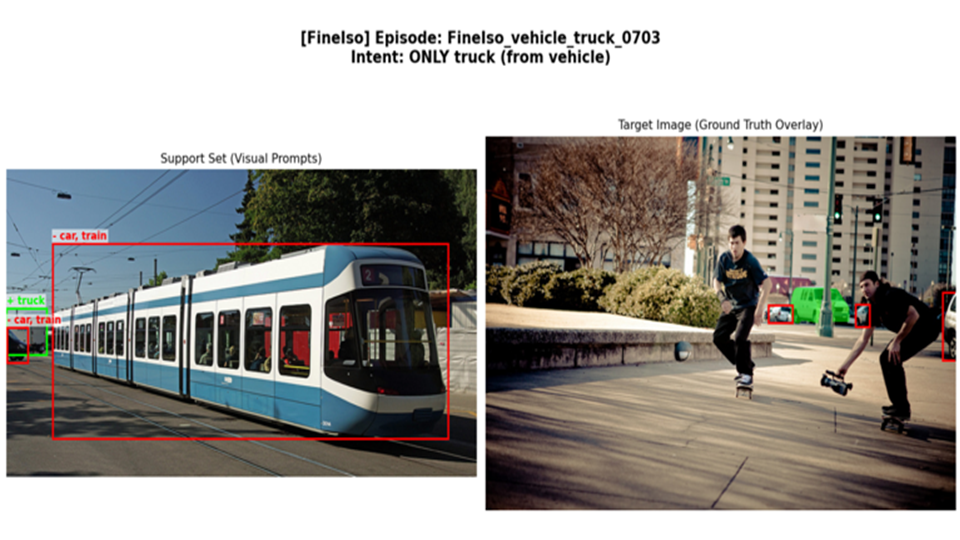}
\caption{
\textbf{Example FineIso episode.} The intent is to isolate trucks from other vehicles.
}
\label{fig:fineiso}
\end{figure}

FineIso evaluates whether a model can isolate a single fine-grained category from related categories within the same supercategory. Each concept is defined as
\[
\mathcal{Y} = \mathcal{C}_{\mathrm{target}},
\]
where $\mathcal{C}_{\mathrm{target}}$ is one COCO category. For example, an episode may define the intent ``only dogs'' within the animal supercategory. Positive prompts are sampled from the target category, while negative prompts are sampled from other categories in the same supercategory, such as cats, sheep, horses, or cows.

The target mask is the union of all instances of the target category:
\[
M = M(\mathcal{C}_{\mathrm{target}}).
\]
The negative prompts do not define a category to subtract from a broader foreground. Instead, they act as discriminative counterexamples that specify the intended granularity of the concept.

For evaluation, target images are required to contain the target category and at least one same-supercategory distractor. Support images are required to contain at least one positive instance and multiple negative distractors. This makes FineIso particularly challenging for models that treat visually related categories as interchangeable.

\subsection{Union: Function-driven Category Union}

\begin{figure}[t]
\centering
\includegraphics[width=\linewidth]{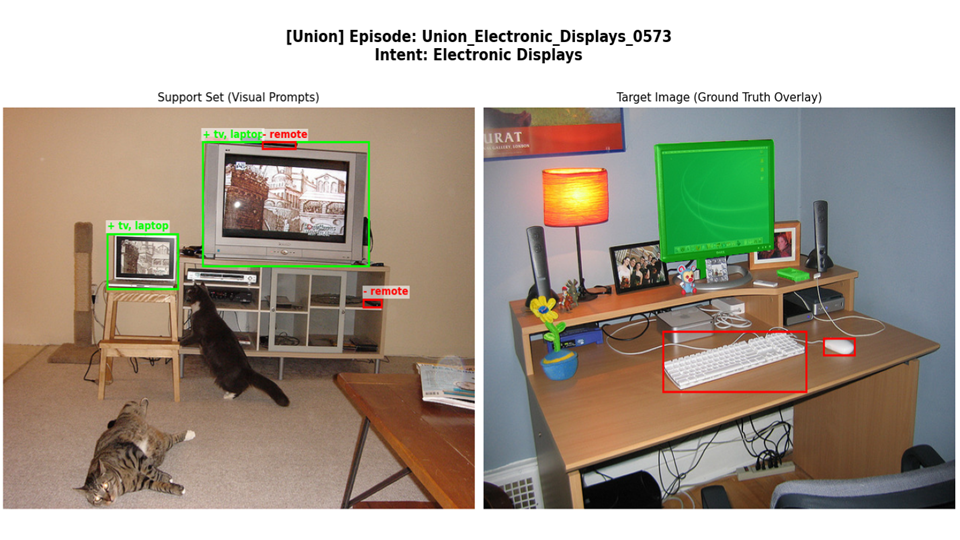}
\caption{
\textbf{Example Union episode.} The intent is to segment electronic displays across related object categories (electronic devices).
}
\label{fig:union}
\end{figure}

Union evaluates whether a model can infer a high-level concept that spans multiple COCO categories. Unlike SuperEx and FineIso, Union is not derived directly from COCO supercategories. Instead, we manually define a fixed set of high-level intent groups. Each group contains positive categories that share a functional or semantic relation and negative categories that serve as related distractors.

Each Union concept is defined as
\[
\mathcal{Y} = \bigcup_{c \in \mathcal{C}_{\mathrm{pos}}} c,
\]
where $\mathcal{C}_{\mathrm{pos}}$ is the manually defined positive category set for a given intent. The target mask is the union of all positive-category instances:
\[
M = \bigcup_{c \in \mathcal{C}_{\mathrm{pos}}} M(c).
\]
For example, the intent ``public transportation'' contains buses and trains as positive categories and cars and motorcycles as negative distractors. Similarly, ``rest furniture'' contains beds and couches as positives and chairs and dining tables as negatives.

The negative prompts in Union are not subtracted from a base mask. Instead, they define the boundary of the high-level concept by showing related categories that should not be included. This setting tests whether a model can infer an abstract group from positive exemplars rather than simply segmenting all objects from a broad COCO supercategory. Table~\ref{tab:supp_union_examples} lists representative Union concepts. The full implementation uses a fixed set of such groups across all experiments.

\begin{table}[t]
\centering
\caption{\textbf{Representative examples of manually defined Union concepts.}}
\label{tab:supp_union_examples}
\scriptsize
\begin{tabularx}{\linewidth}{@{}lXX@{}}
\toprule
Intent & Positive categories & Negative categories \\
\midrule
Public Transportation & bus, train & car, motorcycle \\
Human Powered Mobility & bicycle, skateboard & car, motorcycle \\
Rest Furniture & bed, couch & chair, dining table \\
Domestic Pets & dog, cat & cow, horse, zebra \\
Drink Containers & cup, wine glass, bottle & fork, knife \\
Electronic Displays & tv, laptop, cell phone & remote, keyboard, mouse, microwave \\
Cutlery & fork, knife, spoon & bowl, cup, wine glass, dining table \\
\bottomrule
\end{tabularx}
\end{table}

\subsection{AttrIso: Attribute Isolation within an Object Group}

\begin{figure}[t]
\centering
\includegraphics[width=\linewidth]{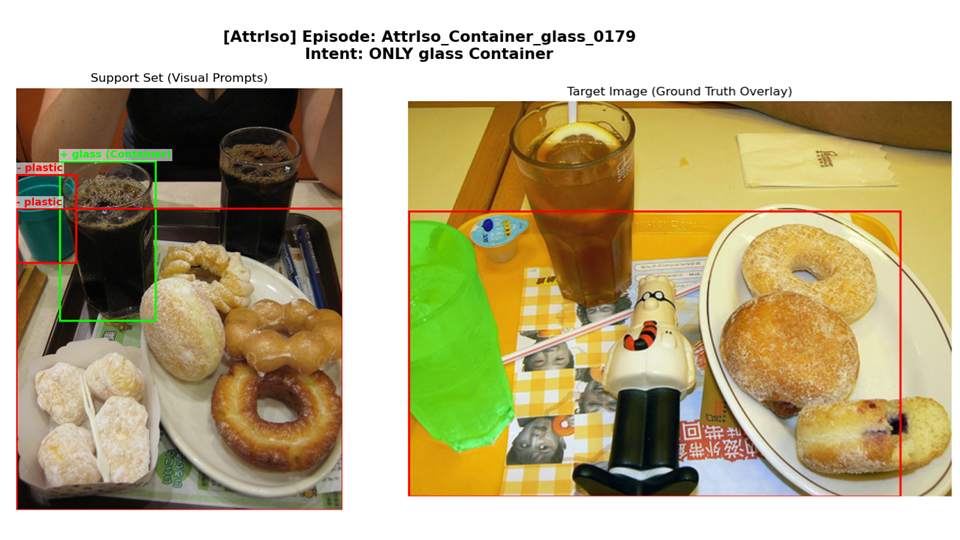}
\caption{
\textbf{Example AttrIso episode.} The intent is to segment glass containers while excluding containers with other attributes.
}
\label{fig:attriso}
\end{figure}

AttrIso evaluates whether a model can isolate objects with a specified visual attribute within a fixed object group. Unlike FineIso, where the concept is defined only by object category, AttrIso defines the target by combining an object group and an attribute:
\[
\mathcal{Y} = \{x \mid x \in \mathcal{G},\ a_{\mathrm{target}} \in A(x)\},
\]
where $\mathcal{G}$ is the target object group, $A(x)$ is the set of PACO attributes assigned to instance $x$, and $a_{\mathrm{target}}$ is the target color or material attribute.

For example, an AttrIso episode may define the intent ``glass containers.'' Positive prompts contain container instances with the target attribute, such as glass containers. Negative prompts contain instances from the same object group with different attributes from the same attribute family, such as plastic containers. This design forces the model to isolate the target attribute while keeping the object group fixed.

The target mask is generated as
\[
M = \bigcup_{x \in \mathcal{G}}
\mathbf{1}[a_{\mathrm{target}} \in A(x)] M(x).
\]

This mask includes only instances that match both the target category and the target attribute. Instances from the same category but with different attributes are excluded, and instances from other categories are not part of the target even if they share the same attribute.

For evaluation, target images are required to contain at least one positive instance and at least one same-category attribute distractor. Support images are required to contain at least one positive prompt and multiple negative prompts from the same object category. We additionally require the negative prompts to contain multiple distractor attributes whenever possible, so that the model cannot solve the episode by suppressing only one specific negative instance.

\subsection{AttrUnion: Cross-category Attribute Union}

\begin{figure}[t]
\centering
\includegraphics[width=\linewidth]{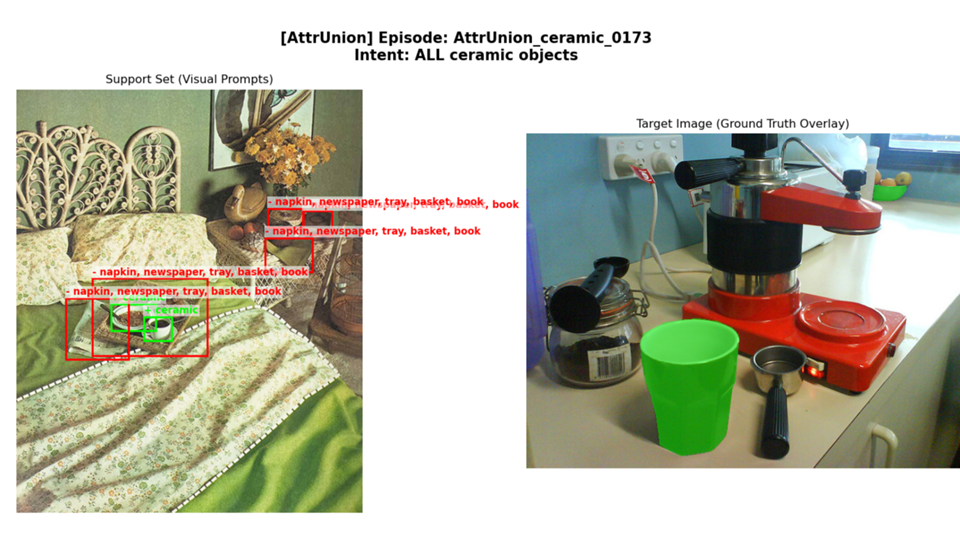}
\caption{
\textbf{Example AttrUnion episode.} The intent is to segment all ceramic objects.
}
\label{fig:attrunion}
\end{figure}

AttrUnion evaluates whether a model can segment all objects that share a specified visual attribute across categories. Each concept is defined as
\[
\mathcal{Y} = \{x \mid a_{\mathrm{target}} \in A(x)\},
\]
where $a_{\mathrm{target}}$ is a target color or material attribute. Unlike AttrIso, AttrUnion does not restrict the target to a semantic group. The target mask includes all valid object-level instances with the target attribute:
\[
M = \bigcup_x \mathbf{1}[a_{\mathrm{target}} \in A(x)] M(x).
\]

For example, an AttrUnion episode may ask the model to segment all wooden objects or all red objects, regardless of their object categories. Positive prompts are sampled from multiple categories that share the target attribute. Negative prompts are sampled from objects that do not have the target attribute but have other attributes from the same attribute family. This prevents the model from solving the task by using object category alone.

For evaluation, we require the target image to contain the target attribute across at least two object categories whenever possible. Support images are also required to contain positive exemplars from multiple categories and negative exemplars with different attributes. This makes AttrUnion a direct test of category-invariant attribute reasoning.

\subsection{AttrEx: Attribute Exclusion within a Semantic Group}

\begin{figure}[t]
\centering
\includegraphics[width=\linewidth]{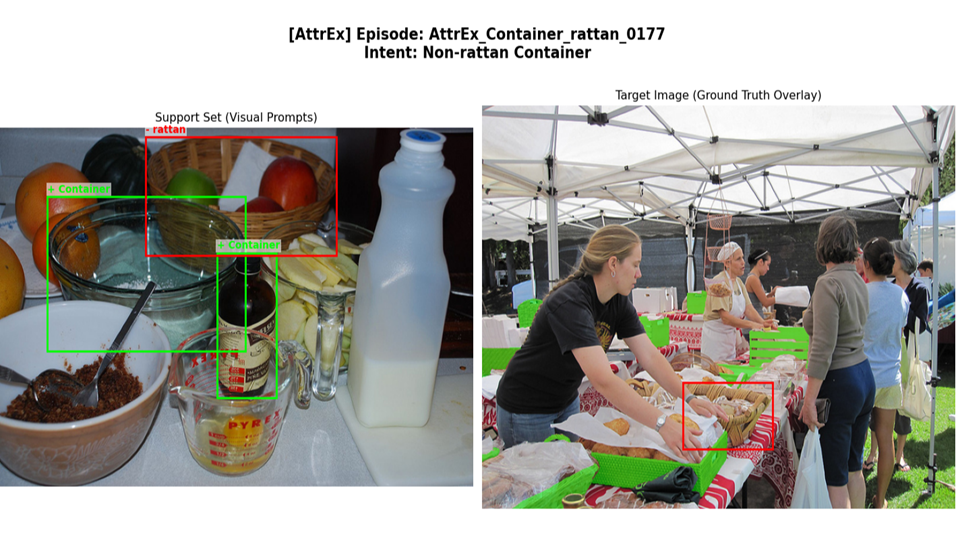}
\caption{
\textbf{Example AttrEx episode.} The intent is to segment non-rattan containers while excluding rattan containers.
}
\label{fig:attrex}
\end{figure}

AttrEx evaluates whether a model can segment objects in a semantic group while excluding a specified visual attribute. Each concept is defined as
\[
\mathcal{Y} = \{x \mid x \in \mathcal{G},\ a_{\mathrm{excluded}} \notin A(x)\},
\]
where $\mathcal{G}$ is a semantic object group and $a_{\mathrm{excluded}}$ is the excluded color or material attribute. For example, an episode may define the intent ``non-red vehicles'' or ``non-wooden furniture.''

The target mask is generated by taking all valid instances in the semantic group and subtracting instances that possess the excluded attribute:
\[
M = M(\mathcal{G}) \setminus 
\bigcup_{x \in \mathcal{G}} \mathbf{1}[a_{\mathrm{excluded}} \in A(x)] M(x).
\]
Positive prompts contain instances from the same semantic group with non-excluded attributes. Negative prompts contain instances with the excluded attribute. This construction requires the model to interpret negative prompts as an attribute-level exclusion rather than a category-level exclusion.

For training, target images are required to contain at least one valid non-excluded foreground instance, while support images must contain both positive and negative exemplars. For evaluation, target and support images are constructed more strictly to include explicit excluded-attribute distractors. We also require the positive prompts to provide diversity either across object categories or across non-excluded attributes. This prevents the task from collapsing into a single-instance matching problem.

\subsection{CASE-NegStress}

CASE-NegStress is a dedicated 1-shot diagnostic subset that fixes the query image, target mask, positive support, and candidate negative pool while varying only the number of negative prompts. The 0-Neg condition denotes positive-only prompting on this same NegStress subset rather than the full CASE. The purpose is not to define a new task family, but to stress-test whether the model's predictions change appropriately when the negative evidence increases. A robust positive-negative visual prompt segmentation model should preserve regions supported by the positive exemplars while suppressing regions that match the negative exemplars. In contrast, a model that ignores negative prompts will produce similar predictions under the original CASE setting and the NegStress setting. Therefore, CASE-NegStress helps distinguish genuine negative-prompt reasoning from positive-only support matching.

\section{Implementation Details}

\subsection{Shared-canvas Formulation}

CPA follows the shared-canvas formulation used in the main paper. Given a $K$-shot support set 
$\mathcal{S}=\{(I_i^{s},\mathcal{B}_i^{+},\mathcal{B}_i^{-})\}_{i=1}^{K}$ 
and a query image $I^{q}$, we compose them into a single RGB canvas $I^{c}$ of size 
$1008 \times 1008$. The positive and negative support boxes are transformed from their original image coordinates into the coordinate system of $I^{c}$ and are used as SAM 3 box prompts. The model predicts composite canvas logits $Z^{c}$, from which we crop the query region to obtain $Z^{q}$ for loss computation and evaluation.

For 1-shot episodes, we use a vertical support-query layout. The support image occupies the top half of the canvas, and the query image occupies the bottom half, where each region is represented as $[x,y,w,h]$:
\[
\Omega^{1\text{-shot}}_{\mathrm{sup}} = [0,0,1008,504],
\qquad
\Omega^{1\text{-shot}}_{\mathrm{qry}} = [0,504,1008,504].
\]

\begin{figure}[t]
\centering
\includegraphics[width=\linewidth]{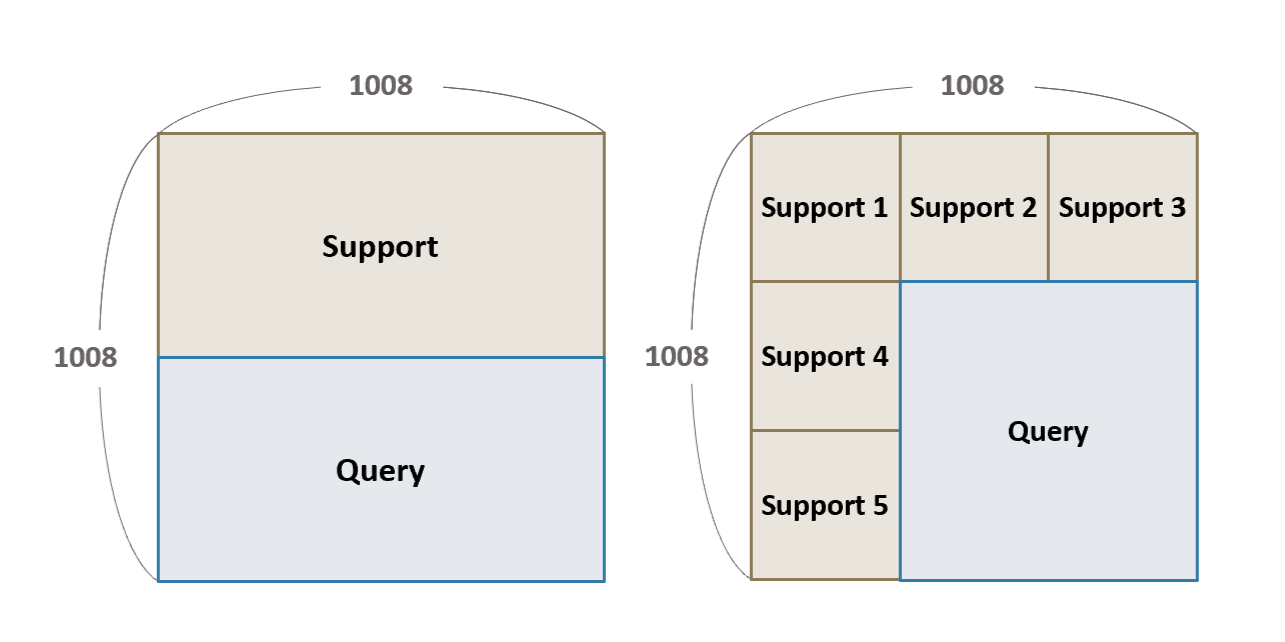}
\caption{\textbf{Shared-canvas layouts used by CPA.} Left: the 1-shot setting uses a vertical layout, where the support image occupies the top half of the $1008 \times 1008$ canvas and the query image occupies the bottom half. Right: the 5-shot setting uses an L-shaped layout, where three support images are placed in the top row, two support images are placed in the left column, and the query image occupies the bottom-right region.}
\label{fig:supp_canvas_layout}
\end{figure}

For 5-shot episodes, we use an L-shaped layout. The first three support images are placed in the top row, the fourth and fifth support images are placed in the left column, and the query image is placed in the bottom-right region. With canvas size $1008$, the support thickness is
\[
s=\mathrm{round}(0.33 \times 1008)=333,
\]
and the query region size is
\[
q=1008-s=675.
\]
Therefore, the 5-shot query region is
\[
\Omega^{5\text{-shot}}_{\mathrm{qry}}=[333,333,675,675].
\]
The top support row consists of three regions of size $336 \times 333$, and the left support column consists of two regions of sizes $333 \times 337$ and $333 \times 338$, respectively. This layout keeps all five support examples visible while preserving a large square query region. Figure~\ref{fig:supp_canvas_layout} illustrates both layouts. 
 
 All support boxes are converted into normalized canvas coordinates in $[c_x,c_y,w,h]$ format. Under positive-negative prompting, both $\mathcal{B}^{+}$ and $\mathcal{B}^{-}$ are provided to the model. Under positive-only prompting, only $\mathcal{B}^{+}$ is retained.

\subsection{LoRA Configuration}

Let $f$ denote the original frozen SAM 3 model and $f_{\theta}$ denote the CPA-adapted model. We keep all pretrained SAM 3 parameters frozen and optimize only the LoRA parameters $\theta$. LoRA adapters are inserted into transformer encoder and decoder layers. In particular, we apply LoRA to modules whose names end with 
\texttt{q\_proj}, \texttt{k\_proj}, \texttt{v\_proj}, \texttt{out\_proj}, \texttt{linear1}, or \texttt{linear2}. 

Unless otherwise specified, we use LoRA rank $r=16$, scaling factor $\alpha=32$, dropout $0.05$, and no bias adaptation. This results in 1.13M trainable parameters. The original SAM 3 parameters remain frozen throughout training, so $f_{\theta}$ differs from $f$ only through the inserted LoRA updates.

\subsection{Training Objective}

CPA is optimized under two prompting conditions, following the notation in the main paper. In the positive-negative condition, the adapted model $f_{\theta}$ receives both positive and negative boxes and predicts query logits $Z^{q}_{\theta,\pm}$:
\[
\mathcal{B}^{+},\mathcal{B}^{-}
\to
f_{\theta}
\to
Z^{q}_{\theta,\pm}.
\]
The prediction is supervised by the CASE ground-truth query mask $M^{q}$:
\[
\mathcal{L}_{\mathrm{svps}}^{\pm}
=
\mathcal{L}_{\mathrm{mask}}
\left(
Z^{q}_{\theta,\pm}, M^{q}
\right).
\]

In the positive-only preservation condition, the adapted model $f_{\theta}$ and the original frozen model $f$ are evaluated under the same positive-only input. The adapted model produces $Z^{q}_{\theta,+}$, while the frozen model produces $Z^{q}_{f,+}$:
\[
\mathcal{B}^{+}
\to
f_{\theta}
\to
Z^{q}_{\theta,+},
\qquad
\mathcal{B}^{+}
\to
f
\to
Z^{q}_{f,+}.
\]
The preservation loss penalizes the discrepancy between their probability maps:
\[
\mathcal{L}_{\mathrm{pres}}^{+}
=
\left\|
\sigma(Z^{q}_{\theta,+})
-
\sigma(Z^{q}_{f,+})
\right\|_2^2,
\]
where $\sigma(\cdot)$ denotes the sigmoid function.

The final training objective is
\[
\mathcal{L}
=
\mathcal{L}_{\mathrm{svps}}^{\pm}
+
\lambda_{\mathrm{pres}}
\mathcal{L}_{\mathrm{pres}}^{+}.
\]
We set $\lambda_{\mathrm{pres}}=1.0$ unless otherwise specified.

In implementation, $\mathcal{L}_{\mathrm{mask}}$ combines sigmoid focal loss and Dice loss:
\[
\mathcal{L}_{\mathrm{mask}}
=
20 \cdot \mathcal{L}_{\mathrm{focal}}
+
\mathcal{L}_{\mathrm{dice}},
\]
where the sigmoid focal loss uses $\alpha=0.25$ and $\gamma=2.0$. When SAM 3 returns multiple mask candidates, we aggregate the candidate logits by taking their element-wise maximum before computing the losses.

\subsection{Optimization Details}

We train CPA on the mixed CASE training set. The dataloader shuffles candidate training episodes, and support examples are sampled according to the requested shot setting. For 1-shot training, one support image is placed in the vertical layout. For 5-shot training, five support images are placed in the L-shaped layout.

We use AdamW with learning rate $1\times10^{-4}$ and weight decay $1\times10^{-4}$. The default training configuration uses a batch size of $1$, two dataloader workers, $10$ epochs, a random seed of $123$, and mixed precision with bfloat16 autocasting on CUDA. Gradients are clipped with a maximum norm of $1.0$. LoRA checkpoints are saved after each epoch.

\begin{table}[t]
\centering
\caption{\textbf{CPA training and optimization settings.}}
\label{tab:supp_training_settings}
\scriptsize
\begin{tabular}{@{}ll@{}}
\toprule
Component & Setting \\
\midrule
Base model & SAM 3 \\
Frozen model notation & $f$ \\
Adapted model notation & $f_{\theta}$ \\
Input canvas size & $1008 \times 1008$ \\
1-shot layout & Vertical support-query split \\
5-shot layout & L-shaped support-query canvas \\
Prompt type & Support-image bounding boxes \\
Positive-negative condition & $\mathcal{B}^{+}$ and $\mathcal{B}^{-}$ \\
Positive-only condition & $\mathcal{B}^{+}$ only \\
Trainable modules & LoRA adapters \\
Frozen modules & All pretrained SAM 3 weights \\
LoRA target modules & \texttt{q\_proj}, \texttt{k\_proj}, \texttt{v\_proj}, \texttt{out\_proj}, \texttt{linear1}, \texttt{linear2} \\
LoRA rank & $r=16$ \\
LoRA scaling factor & $\alpha=32$ \\
LoRA dropout & $0.05$ \\
Trainable parameters & 1.13M \\
Optimizer & AdamW \\
Learning rate & $1\times10^{-4}$ \\
Weight decay & $1\times10^{-4}$ \\
Random seed & $123$ \\
Mixed precision & bfloat16 autocast on CUDA \\
Gradient clipping & Max norm $1.0$ \\
Preservation weight & $\lambda_{\mathrm{pres}}=1.0$ \\
\bottomrule
\end{tabular}
\end{table}

\subsection{Inference and Evaluation Protocol}

At inference time, we use the same shared-canvas formulation as in training. Given $\mathcal{S}$ and $I^{q}$, we compose the canvas $I^{c}$, transform the support boxes into the canvas coordinate system, and apply $f_{\theta}$ to obtain composite logits $Z^{c}$. We crop the query region to obtain $Z^{q}$ and derive the binary query prediction as
\[
\hat{M}^{q}
=
\mathbf{1}[\sigma(Z^{q})>\tau],
\]
where $\sigma(\cdot)$ denotes the sigmoid function and $\tau$ is the mask threshold. We use $\tau=0.5$ for all reported results. We then resize the query prediction to the original query resolution and compute mIoU against the ground-truth query mask $M^{q}$.

For CASE, the fixed evaluation episode provides the support images and the positive/negative support boxes. We use the vertical layout for 1-shot evaluation and the L-shaped layout for 5-shot evaluation. For standard FSS benchmarks, we follow the conventional positive-only setting: support masks are converted into tight bounding boxes and used as positive visual prompts, and no negative prompts are provided. For COCO-$20^i$, the CASE-trained CPA checkpoints are evaluated directly on all four folds without COCO-$20^i$ retraining or fold-specific tuning. This evaluation measures CPA's positive-only FSS behavior after adaptation to CASE rather than assuming that the original behavior is fully preserved.

\subsection{Baseline Adaptation}
\begin{table*}[t]
\centering
\caption{\textbf{Split-wise FSSAM results on CASE.} FSSAM is evaluated using the publicly released split-specific checkpoints. The main paper reports the average over four splits.}
\label{tab:supp_fssam_splits}
\scriptsize
\begin{tabular*}{\textwidth}{@{\extracolsep{\fill}} l l c c c c c @{}}
\toprule
Shot & Task & Split 0 & Split 1 & Split 2 & Split 3 & Avg. \\
\midrule
\multirow{6}{*}{1-shot}
& FineIso   & 40.75 & 37.74 & 38.25 & 39.15 & 38.97 \\
& SuperEx   & 27.88 & 28.80 & 29.19 & 28.35 & 28.55 \\
& Union     & 36.89 & 34.92 & 35.96 & 35.80 & 35.89 \\
& AttrIso   & 18.69 & 13.90 & 15.04 & 15.84 & 15.87 \\
& AttrUnion & 15.04 & 15.25 & 13.90 & 15.13 & 14.83 \\
& AttrEx    & 17.25 & 17.16 & 17.31 & 17.40 & 17.28 \\
\midrule
\multirow{6}{*}{5-shot}
& FineIso   & 54.84 & 54.52 & 55.72 & 55.45 & 55.13 \\
& SuperEx   & 35.36 & 38.74 & 37.07 & 37.26 & 37.11 \\
& Union     & 41.85 & 46.98 & 44.26 & 44.94 & 44.51 \\
& AttrIso   & 19.17 & 16.40 & 18.18 & 17.71 & 17.87 \\
& AttrUnion & 19.36 & 21.52 & 20.09 & 21.49 & 20.61 \\
& AttrEx    & 18.10 & 17.46 & 17.57 & 17.90 & 17.76 \\
\bottomrule
\end{tabular*}
\end{table*}

For SAM-based FSS baselines that do not natively support negative visual exemplars, we adapt the CASE inputs to their supported prompt formats. GF-SAM and FSSAM use positive support masks derived from the positive boxes, while SANSA uses its native positive-box prompt interface. Since these methods were designed for conventional positive-only FSS, their CASE results should be interpreted as positive-only reference baselines rather than fully specified SVPS.

For the main-paper comparison, we use the publicly released checkpoints provided by the authors' official repositories and do not fine-tune these baselines on CASE. A separate same-data SANSA control is reported below to isolate the effect of in-domain CASE exposure. For methods with a single released checkpoint, we report the result obtained from that checkpoint. For FSSAM, the official release provides split-specific checkpoints. We therefore evaluate FSSAM on each of the four splits and report the average across splits in the main paper. The reported FSSAM average is computed at the metric level over the four split-specific evaluations, not by ensembling predictions from multiple checkpoints. The split-wise FSSAM results are shown in Table~\ref{tab:supp_fssam_splits}.

\begin{table*}[t]
\centering
\caption{\textbf{Full aggregated CASE results across concept dimensions.} Category and Attribute correspond to the source concept type, while Exclusion, Isolation, and Union correspond to the compositional operation required by the task.}
\label{tab:supp_case_summary_full}
\scriptsize
\begin{tabular*}{\textwidth}{@{\extracolsep{\fill}} l c c c c c c c @{}}
\toprule
Method & Prompt & Category & Attribute & Exclusion & Isolation & Union & \textbf{Mean} \\
\midrule
\multicolumn{8}{c}{\textbf{1-shot}} \\
\midrule
GF-SAM~\cite{zhang2024gfsam} & Pos Only & 31.24 & 14.81 & 20.80 & 26.52 & 21.75 & 23.03 \\
FSSAM~\cite{xu2025fssam} & Pos Only & 34.47 & 15.99 & 22.92 & 27.42 & 25.36 & 25.23 \\
SANSA~\cite{cuttano2025sansa} & Pos Only & 36.81 & 18.40 & 21.30 & 38.07 & 23.45 & 27.61 \\
FSS-SAM3~\cite{tsai2026fsssam3} & Pos Only & 42.82 & 18.40 & 24.43 & 38.03 & 29.38 & 30.61 \\
FSS-SAM3~\cite{tsai2026fsssam3} & Pos+Neg & 25.72 & 11.52 & 21.64 & 13.38 & 20.85 & 18.62 \\
\midrule
CPA (Ours) & Pos Only & 37.26 & 19.70 & 23.67 & 35.55 & 26.24 & 28.48 \\
\textbf{CPA (Ours)} & \textbf{Pos+Neg} & \textbf{48.63} & \textbf{23.04} & \textbf{29.55} & \textbf{42.72} & \textbf{35.23} & \textbf{35.83} \\
\midrule
\multicolumn{8}{c}{\textbf{5-shot}} \\
\midrule
GF-SAM~\cite{zhang2024gfsam} & Pos Only & 43.47 & 21.18 & 28.24 & 36.10 & 32.64 & 32.32 \\
FSSAM~\cite{xu2025fssam} & Pos Only & 45.58 & 18.75 & 27.43 & 36.50 & 32.56 & 32.17 \\
SANSA~\cite{cuttano2025sansa} & Pos Only & 44.27 & 22.52 & 26.41 & 43.40 & 30.38 & 33.40 \\
FSS-SAM3~\cite{tsai2026fsssam3} & Pos Only & 51.72 & 24.48 & 31.09 & 43.80 & 39.43 & 38.10 \\
FSS-SAM3~\cite{tsai2026fsssam3} & Pos+Neg & 46.03 & 22.26 & 30.25 & 36.16 & 36.03 & 34.14 \\
\midrule
CPA (Ours) & Pos Only & 52.44 & 26.16 & 35.84 & 44.03 & 38.04 & 39.30 \\
\textbf{CPA (Ours)} & \textbf{Pos+Neg} & \textbf{57.15} & \textbf{27.14} & \textbf{37.74} & \textbf{46.82} & \textbf{41.88} & \textbf{42.15} \\
\bottomrule
\end{tabular*}
\end{table*}

\section{Additional Experimental and Qualitative Results}

\subsection{Full Concept-dimension Breakdown on CASE}
\cref{tab:supp_case_summary_full} provides the full concept-dimension breakdown corresponding to the aggregated results in the main paper. Category and Attribute summarize the source type of the target concept. Exclusion, Isolation, and Union summarize the operation required to define the target set.

The results show that negative visual exemplars are useful beyond category-level matching. CPA improves performance on the exclusion and isolation dimensions, indicating that negative prompts help reject explicitly excluded categories or attributes and sharpen distinctions among related objects. Union-style concepts also benefit, suggesting that negative exemplars help constrain broad or compositional target groups. Attribute-level dimensions remain more challenging, consistent with the observation that color and material cues are often subtle and context-dependent.

\subsection{Additional Quantitative Analyses}

\subsubsection{Precision--recall behavior under positive and negative prompting.}

\begin{table}[t]
\centering
\caption{\textbf{1-shot CASE precision--recall analysis.} All methods use the same evaluation episodes and $\tau=0.5$.}
\label{tab:supp_prf}
\scriptsize
\setlength{\tabcolsep}{4.5pt}
\begin{tabular}{@{}llcccc@{}}
\toprule
Model & Prompt & mIoU & Precision & Recall & F1 \\
\midrule
FSS-SAM3 & Pos & 30.61 & 64.10 & 36.21 & 46.28 \\
CPA (Ours) & Pos & 28.48 & 35.12 & \textbf{58.02} & 43.75 \\
FSS-SAM3 & Pos+Neg & 18.62 & \textbf{68.29} & 20.66 & 31.72 \\
CPA (Ours) & Pos+Neg & \textbf{35.83} & 63.32 & 55.72 & \textbf{59.28} \\
\bottomrule
\end{tabular}
\end{table}

CPA is adapted to infer concepts jointly from positive and negative exemplars. Consequently, under positive-only inference it retains more candidate regions than FSS-SAM3, yielding higher recall but lower precision. When negative exemplars are provided, CPA recovers precision while retaining substantially higher recall, producing higher F1 and mIoU than FSS-SAM3 under Pos+Neg prompting.

\subsubsection{Same-data training control.}
To separate the benefit of CASE training data from the ability to use negative exemplars, we fine-tune SANSA on the same CASE training set while retaining its native positive-only prompting interface, which does not support negative visual exemplars. In-domain training improves SANSA, but CPA remains $+5.82/+5.88$ mIoU higher in 1-/5-shot, showing that CASE exposure alone is insufficient to explain CPA's gains.

\begin{table}[t]
\centering
\caption{\textbf{In-domain training control on CASE} (mIoU).}
\label{tab:supp_training_control}
\scriptsize
\setlength{\tabcolsep}{5pt}
\begin{tabular}{@{}lcccc@{}}
\toprule
Method & CASE FT & Prompt & 1-shot & 5-shot \\
\midrule
SANSA & No & Pos & 27.61 & 33.40 \\
SANSA & Yes & Pos & 30.01 & 36.27 \\
CPA (Ours) & Yes & Pos+Neg & \textbf{35.83} & \textbf{42.15} \\
\bottomrule
\end{tabular}
\end{table}

\subsubsection{SAM 3 text-prompt controls.}
SAM 3 supports text prompts, but exclusion intent is not reliably expressed by the tested text formulations. On SuperEx, the reviewer-inspired exclusion prompt ``[supercategory] except [excluded]'' obtains 5.79 mIoU as a text-only prompt. Combining text with visual prompts remains below the corresponding visual-only FSS-SAM3 results in the main paper (28.80/39.07 mIoU in 1-/5-shot).

\begin{table*}[t]
\centering
\caption{\textbf{SAM 3 text controls on SuperEx} (mIoU). The two blocks are separate controls and are not row-wise paired.}
\label{tab:supp_text_controls}
\scriptsize
\begin{minipage}[t]{0.47\textwidth}
\centering
\begin{tabular}{@{}lcc@{}}
\toprule
Prompt Type & 1-shot & 5-shot \\
\midrule
Text only & 5.79 & 5.79 \\
Text + Pos Visual & 14.26 & 26.03 \\
Text + Pos+Neg Visual & 13.58 & 28.05 \\
\bottomrule
\end{tabular}
\end{minipage}
\hfill
\begin{minipage}[t]{0.49\textwidth}
\centering
\begin{tabular}{@{}lc@{}}
\toprule
Text Formulation & SuperEx \\
\midrule
``[supercategory]'' & 26.28 \\
``[supercategory] except [excluded]'' & 5.79 \\
``[supercategory] not [excluded]'' & 2.60 \\
\bottomrule
\end{tabular}
\end{minipage}
\end{table*}

\subsubsection{Human validation of CASE.}
To quantify benchmark quality and the PACO ambiguity discussed in the failure analysis, two annotators independently reviewed 300 fixed 1-shot evaluation episodes (50 per task family) using four 1--5 criteria: intent quality, positive-exemplar consistency, negative-exemplar consistency, and target-mask consistency. The four mean scores are 4.22/4.38/4.54/4.10, with medians of 5 for all criteria. Across criteria, 84.7--86.7\% of paired ratings are within one point, with linear-weighted $\kappa=0.42$--$0.57$. PACO-derived attribute episodes are more ambiguous than COCO-derived category episodes, especially for target-mask consistency (3.57 vs. 4.63), quantitatively supporting the limitation discussed below.

\begin{table}[t]
\centering
\caption{\textbf{Human validation of CASE.} Agreement is computed over 300 paired episode ratings.}
\label{tab:supp_human_audit}
\scriptsize
\setlength{\tabcolsep}{4pt}
\begin{tabular}{@{}lccc@{}}
\toprule
Criterion & Mean & Within $\pm1$ & Linear $\kappa$ \\
\midrule
Intent quality & 4.22 & 84.7\% & 0.453 \\
Positive consistency & 4.38 & 85.3\% & 0.490 \\
Negative consistency & 4.54 & 86.0\% & 0.417 \\
Target consistency & 4.10 & 86.7\% & 0.570 \\
\bottomrule
\end{tabular}
\end{table}

\subsection{Additional Qualitative Results}

This section provides additional qualitative comparisons complementing the main paper. Unless otherwise noted, green boxes on support images denote positive visual prompts and red boxes denote negative visual prompts. On query images, yellow contours denote ground-truth masks and blue overlays denote model predictions. We compare the frozen SAM 3 model $f$ and the CPA-adapted model $f_{\theta}$ under both positive-only and positive-negative prompting. In the support-prompt column, the support regions follow the same shared-canvas layout used for inference, while the query region is grayed out.

\begin{figure*}[t]
\centering
\includegraphics[width=\textwidth]{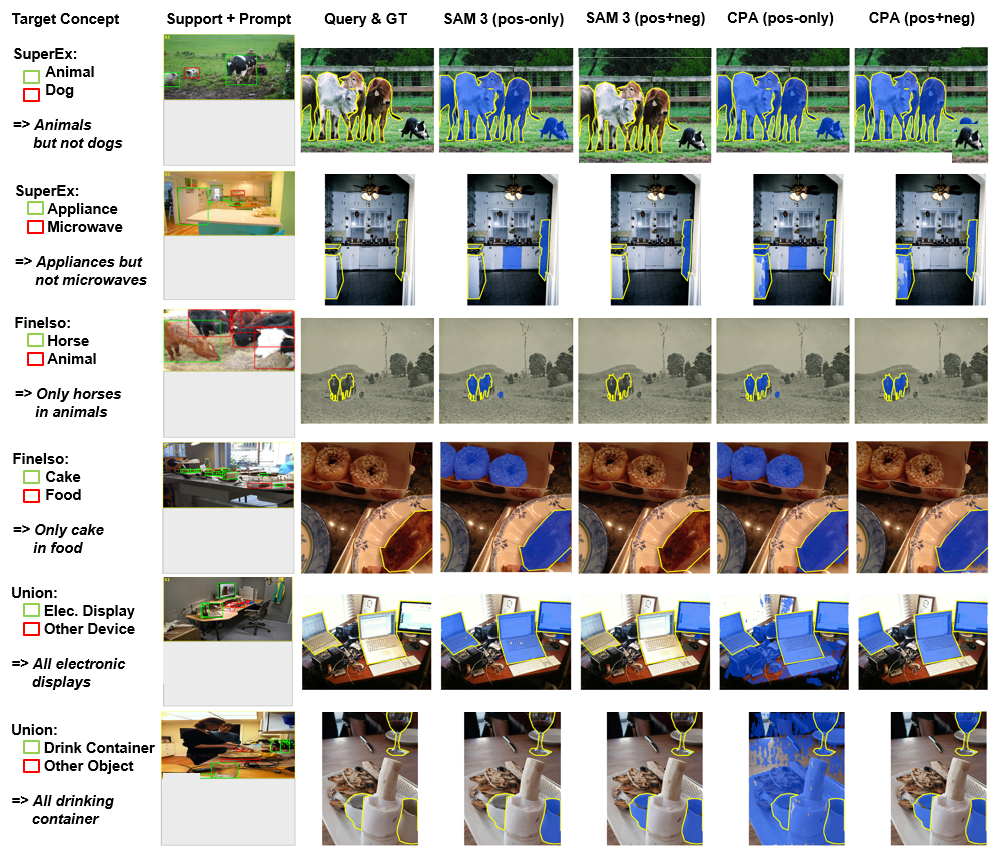}
\caption{\textbf{Additional 1-shot qualitative comparisons on category-level CASE tasks.} The figure includes two examples from each category-level task family: SuperEx, FineIso, and Union. In the support-prompt column, the upper region shows the 1-shot support image with positive and negative box prompts, while the lower gray region indicates that the query image is omitted from the prompt visualization. Frozen SAM 3 with positive-only prompts often follows visual similarity without resolving the intended exclusion, isolation, or union constraint. Directly adding negative prompts to frozen SAM 3 is unreliable, either having little effect or suppressing valid target regions. CPA with positive-only prompts produces cleaner positive-prompt responses, while CPA with positive-negative prompts better matches the concept jointly specified by positive and negative exemplars.}
\label{fig:supp_qual_cat}
\end{figure*}

\begin{figure*}[t]
\centering
\includegraphics[width=\textwidth]{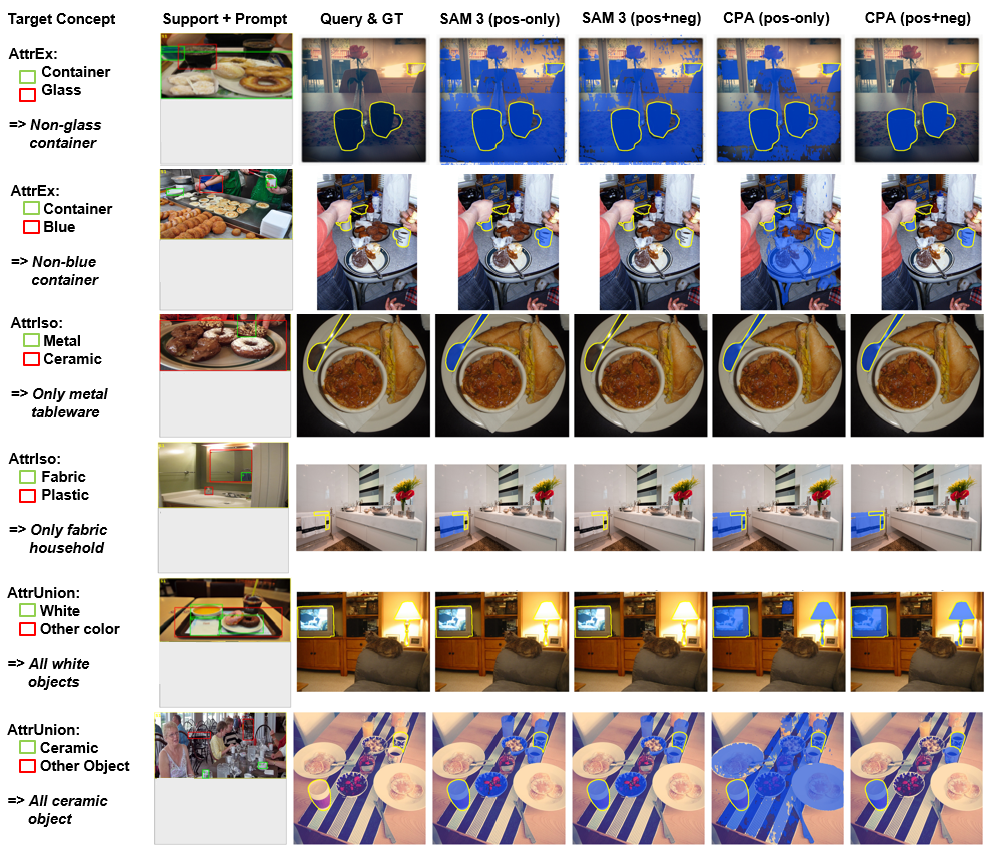}
\caption{\textbf{Additional 1-shot qualitative comparisons on attribute-level CASE tasks.} The figure includes two examples from each attribute-level task family: AttrEx, AttrIso, and AttrUnion. These examples highlight the difficulty of specifying targets by color or material cues rather than object category alone. Frozen SAM 3 tends to rely on category-level or appearance-level similarity and may fail to respect the intended attribute constraint. CPA improves the use of positive and negative attribute exemplars, although attribute-level concepts remain more challenging than category-level concepts due to subtle visual cues and context-dependent material appearance.}
\label{fig:supp_qual_attr}
\end{figure*}

\cref{fig:supp_qual_cat,fig:supp_qual_attr} show representative 1-shot examples across category-level and attribute-level CASE tasks. The examples illustrate that positive-only prompting is often insufficient for specified concepts, since the model may segment visually related distractors together with the intended target or follow object-category priors instead of the specified attribute. Negative prompts are necessary to specify what should be rejected, but frozen SAM 3 does not consistently use them as concept-level constraints. CPA provides more stable predictions by adapting the model to interpret positive and negative visual exemplars jointly.

\begin{figure*}[t]
\centering
\includegraphics[width=\textwidth]{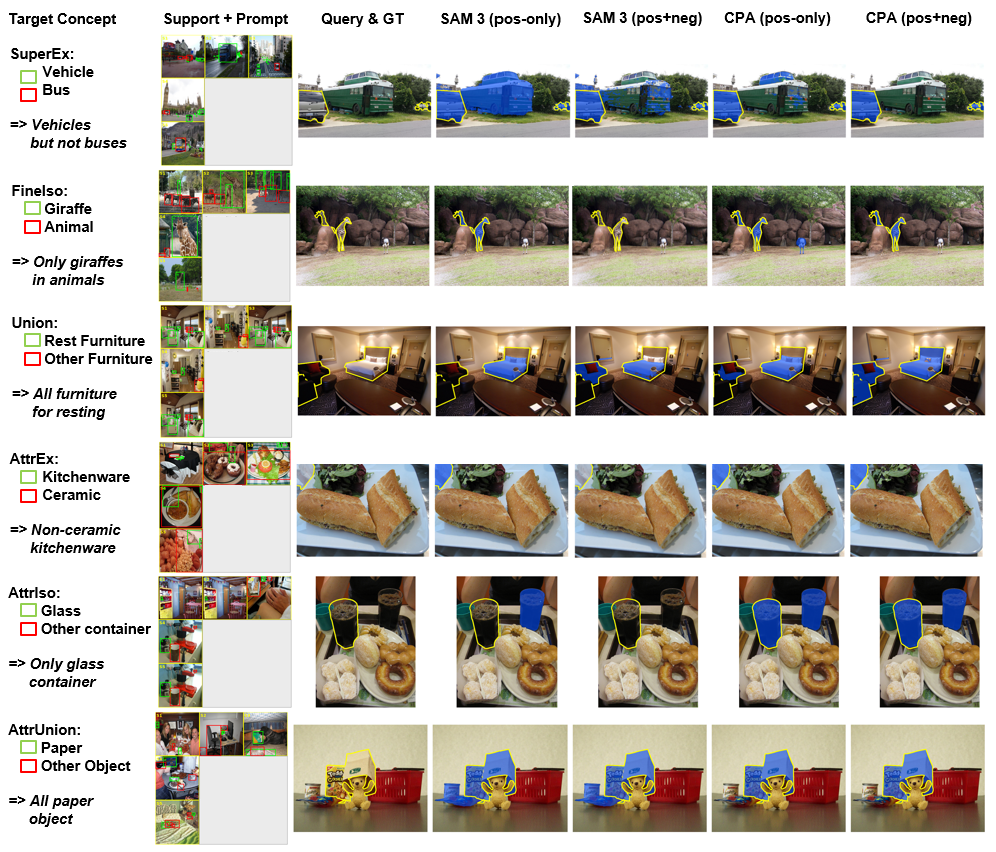}
\caption{\textbf{Additional 5-shot qualitative examples on CASE.} The figure includes one example from each of the six task families. The support-prompt column follows the L-shaped 5-shot shared-canvas layout, where the five support images occupy the top row and left column, and the lower-right gray region indicates that the query image is omitted from the prompt visualization. Multiple support exemplars provide more evidence about intra-concept variation, but the target concept can remain ambiguous without properly using negative exemplars. CPA uses the expanded support set more effectively under positive-negative prompting.}
\label{fig:supp_qual_5shot}
\end{figure*}

\cref{fig:supp_qual_5shot} provides 5-shot qualitative examples. Compared with the 1-shot setting, multiple support images provide broader visual evidence about the intended target concept. However, simply increasing the number of positive exemplars does not fully resolve exclusion, isolation, union, or attribute ambiguity. The positive-negative CPA predictions are generally more consistent with the target masks, indicating that the model benefits from both support diversity and explicit negative evidence.

\section{Failure Cases and Limitations}

\begin{figure*}[t]
\centering
\includegraphics[width=\textwidth]{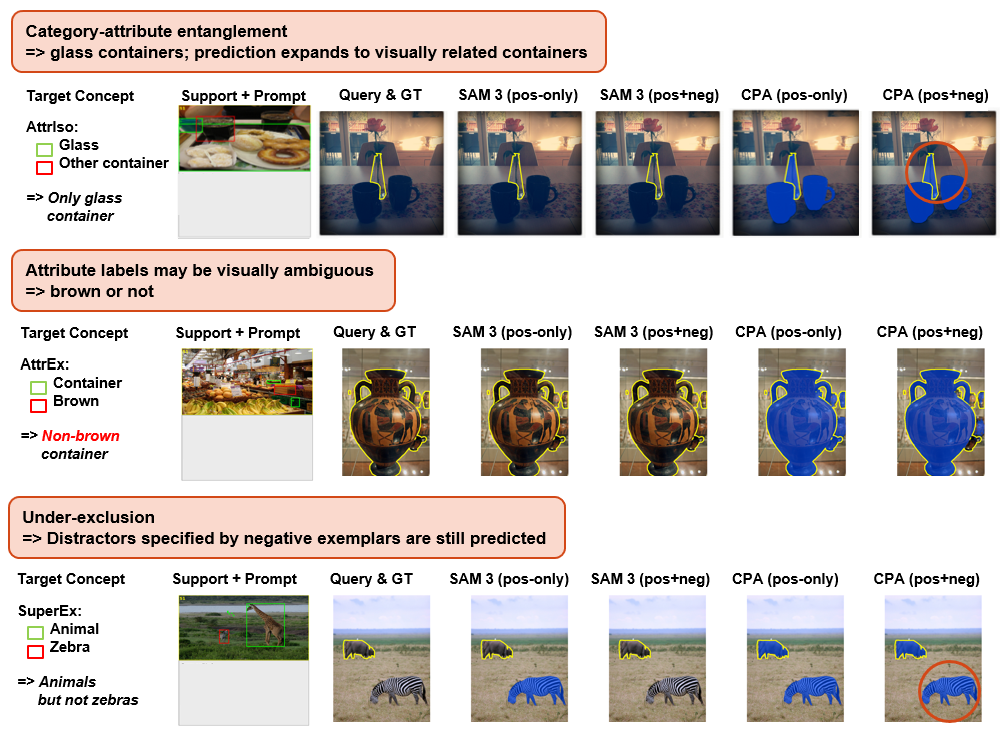}
\caption{
\textbf{Representative failure cases.}
Yellow contours denote ground-truth masks and blue overlays denote model predictions. Red circles highlight erroneous regions. 
Top: in attribute-isolation cases, the adapted model may follow object-category priors and expand from the specified attribute target to visually related containers.
Middle: attribute annotations can be visually ambiguous; for example, the distinction between brown and non-brown containers may depend on lighting, texture, and annotation convention.
Bottom: negative exemplars do not always fully suppress distractors, leading to under-exclusion when excluded categories remain visually salient in the query image.
}
\label{fig:supp_failure_cases}
\end{figure*}

\begin{table*}[t]
\centering
\caption{\textbf{Summary of common failure modes in CASE.}}
\label{tab:supp_failure_modes}
\scriptsize
\begin{tabularx}{\textwidth}{@{}lXX@{}}
\toprule
Failure mode & Description & Typical cause \\
\midrule
Attribute ambiguity
& The model fails to distinguish visually subtle colors or materials, such as glossy plastic versus glass or dark brown versus black.
& PACO attributes can be visually subtle, lighting-dependent, or partially visible. \\

Over-exclusion
& Negative exemplars suppress valid foreground regions together with distractors.
& The negative prompt may share shape, texture, or context with the target concept. \\

Under-exclusion
& Distractor instances matching the negative exemplars remain in the prediction.
& The model relies on positive similarity and does not fully use negative evidence. \\

Category-attribute entanglement
& The model follows object category more strongly than the specified attribute.
& Foundation segmentation models retain object-centric priors from pretraining. \\

Small or occluded objects
& The model misses small, partially visible, or heavily occluded target instances.
& Shared-canvas resizing reduces fine details, and support-query matching becomes less reliable. \\

Ambiguous grouping
& The model inconsistently handles broad union concepts or semantic object groups.
& Some manually defined groups reflect functional relations that are not always visually apparent. \\
\bottomrule
\end{tabularx}
\end{table*}

Although CPA improves the use of positive and negative visual exemplars, several limitations remain. \cref{fig:supp_failure_cases,tab:supp_failure_modes} summarize representative failure modes. Attribute-level tasks are particularly challenging. Color and material labels may depend on lighting, viewpoint, texture, and object state. Some materials are also visually confounded with object categories; for example, glass may be correlated with bottles or cups, while fabric may be correlated with furniture or clothing. As a result, the model may segment objects that match the category prior rather than the specified attribute.

\cref{fig:supp_failure_cases} illustrates three representative limitations. First, attribute-level tasks remain vulnerable to category--attribute entanglement: the model may segment objects that match a strong object-category prior, such as containers, rather than strictly following the specified attribute constraint. Second, some PACO attributes are visually ambiguous, especially for color and material labels affected by lighting, texture, viewpoint, or annotation convention. Third, negative exemplars may still lead to under-exclusion when the excluded category remains visually salient in the query image.

Negative prompts can also introduce failure modes. When positive and negative exemplars are visually similar, the model may over-exclude valid target regions. Conversely, when negative exemplars are visually diverse or weakly related to the query distractors, the model may under-exclude and behave similarly to positive-only FSS. These errors indicate that positive-negative visual prompting requires not only recognizing individual support instances, but also inferring the intended relation between inclusion and exclusion evidence.

CASE itself is limited by the annotation spaces of COCO and PACO. COCO provides reliable category-level instance masks but does not contain fine-grained attributes. PACO provides attribute annotations, but the available color and material labels are incomplete and sometimes ambiguous. Union concepts are manually defined and capture only a subset of possible high-level semantic groupings. Extending CASE to richer attributes, open-vocabulary concepts, and more diverse real-world domains is an important direction for future work.

Finally, CPA is designed as a lightweight and conservative adaptation of SAM 3. The preservation regularization mitigates positive-only FSS degradation, but does not fully preserve the frozen model in the 1-shot setting; it also introduces an adaptation--preservation trade-off. Stronger adaptation may improve CASE performance further, but could degrade the original ability of SAM 3. Exploring stronger architectures or training objectives that improve SVPS without sacrificing positive-only FSS remains future work.

\FloatBarrier 
\end{document}